\documentclass[journal,10pt,onecolumn,draftclsnofoot]{IEEEtran}

\IEEEoverridecommandlockouts
\usepackage{cite}
\usepackage{amsmath,amssymb,amsfonts}
\usepackage{algorithmic}
\usepackage{graphicx}
\usepackage{textcomp}
\usepackage{xcolor}
\usepackage{tikz}
\usepackage{url}
\usepackage{standalone}
\def\BibTeX{{\rm B\kern-.05em{\sc i\kern-.025em b}\kern-.08em
    T\kern-.1667em\lower.7ex\hbox{E}\kern-.125emX}}

\newcommand{\naive}{na\"ive }
\newcommand{\Naive}{Na\"ive }
\newcommand{\algcomment}[1]{\textcolor{teal}{$\triangleleft\quad$\textit{#1}$\quad\triangleright$}\;}
\usepackage{ml_package}

\begin{document}

\title{Calibrating AI Models for Wireless~Communications via Conformal Prediction}

\author{    
    Kfir~M.~Cohen, ~\IEEEmembership{Student Member,~IEEE},
    Sangwoo~Park, ~\IEEEmembership{Member,~IEEE},\\
    Osvaldo~Simeone, ~\IEEEmembership{Fellow,~IEEE},
    and Shlomo~Shamai~(Shitz),~\IEEEmembership{Life~Fellow,~IEEE}
     
    \thanks{Part of this work has been submitted to the 2023 IEEE International Conference on Acoustics, Speech, and Signal Processing (ICASSP 2023).}
    \thanks{The work of Kfir M. Cohen, Sangwoo Park and Osvaldo Simeone has been supported by the European Research Council (ERC) under the European Union’s Horizon 2020 research and innovation programme, grant agreement No. 725731. The work of Osvaldo Simeone has also been supported by an Open Fellowship of the EPSRC with reference EP/W024101/1.}
    \thanks{The work of Shlomo Shamai has been supported by the European Union's Horizon 2020 Research And Innovation Programme, grant agreement No. 694630.}
    \thanks{The authors acknowledge the use of King's Computational Research, Engineering and Technology Environment (CREATE). Retrieved November 21, 2022, from https://doi.org/10.18742/rnvf-m076.}
    \thanks{Kfir M. Cohen, Sangwoo Park, and Osvaldo Simeone are with King's Communication, Learning, \& Information Processing (KCLIP) lab, Department of Engineering, King’s College London, London WC2R 2LS, U.K. (e-mail: kfir.cohen@kcl.ac.uk; sangwoo.park@kcl.ac.uk; osvaldo.simeone@kcl.ac.uk).}
    \thanks{Shlomo Shamai (Shitz) is with the Viterbi Faculty of Electrical and Computing Engineering, Technion—Israel Institute of Technology, Haifa, Israel 3200003 (e-mail: sshlomo@ee.technion.ac.il).}
}

\maketitle

\begin{abstract}
When used in complex engineered systems, such as communication networks, artificial intelligence (AI) models should be not only as accurate as possible, but also well calibrated. A well-calibrated AI model is one that can reliably quantify the uncertainty of its decisions, assigning high confidence levels to decisions that are likely to be correct and low confidence levels to decisions that are likely to be erroneous. This paper investigates the application of conformal prediction as a general framework to obtain AI models that produce decisions with formal calibration guarantees. Conformal prediction transforms probabilistic predictors into set predictors that are guaranteed to contain the correct answer with a probability chosen by the designer. Such formal calibration guarantees hold irrespective of the true, unknown, distribution underlying the generation of the variables of interest, and can be defined in terms of ensemble or time-averaged probabilities. In this paper, conformal prediction is applied for the first time to the design of AI for communication systems in conjunction to both frequentist and Bayesian learning, focusing on demodulation, modulation classification, and channel prediction.
\end{abstract}


\section{Introduction} \label{sec: Introduction}

\subsection{Motivation}
How reliable is your artificial intelligence (AI)-based model? The most common metric to design an AI model and to gauge its performance is the average \emph{accuracy}. However, in applications in which AI decisions are used within a larger system, AI models should not only be as accurate as possible, but they should also be able to reliably quantify the uncertainty of their decisions. As an example, consider an unlicensed link that uses AI tools to predict the best channel to access out of four possible channels. A predictor that assigns the probability vector of $[90\%,2\%,5\%,3\%]$ to the possible channels predicts the same best channel -- the first -- as a predictor that outputs the probability vector $[30\%,20\%,25\%,25\%]$. However, the latter predictor is less certain of its decision, and it may be preferable for the unlicensed link to refrain from accessing the channel when acting on less confident predictions, e.g., to avoid excessive interference to licensed links \cite{wang2010advances,simeone2008spectrum}.

As in the example above, AI models typically report a confidence measure associated with each prediction, which reflects the model's \emph{self-evaluation} of the accuracy of a decision. Notably, neural network models implement \emph{probabilistic predictors} that produce a probability distribution across all possible values of the output variable. 
The self-reported model confidence, however, may not be a reliable measure of the true, unknown, accuracy of a prediction. In such situations, the AI model is said to be poorly calibrated.

As illustrated in the example in Fig.~\ref{fig: fig_tikz_4classes_acc_cal}, accuracy and calibration are distinct criteria, with neither criterion implying the other. It is, for instance, possible to have an accurate predictor that consistently underestimates the accuracy of its decisions, and/or that is overconfident where making incorrect decisions (see fourth column in Fig.~\ref{fig: fig_tikz_4classes_acc_cal}). Conversely, 
one can have inaccurate predictions that estimate correctly their uncertainty (see fifth column in Fig.~\ref{fig: fig_tikz_4classes_acc_cal}).

Deep learning models tend to produce either overconfident decisions \cite{Guo2017Calibration}, or calibration levels that rely on strong assumptions about the ground-truth, unknown, data generation mechanism \cite{masegosa2020learning, morningstar2022pacm, zecchin2022robustpacm,cannon2022investigating,frazier2020model,ridgway2017probably}. This paper investigates the use of \emph{conformal prediction} (CP) \cite{vovk2005algorithmic, zeni2020conformal,angelopoulos2021gentle} as a framework to design provably well-calibrated AI predictors, with \emph{distribution-free} calibration guarantees that do not require making any assumption about the ground-truth data generation mechanism.

\subsection{Conformal Prediction for AI-Based Wireless Systems}

\begin{figure*}[t]
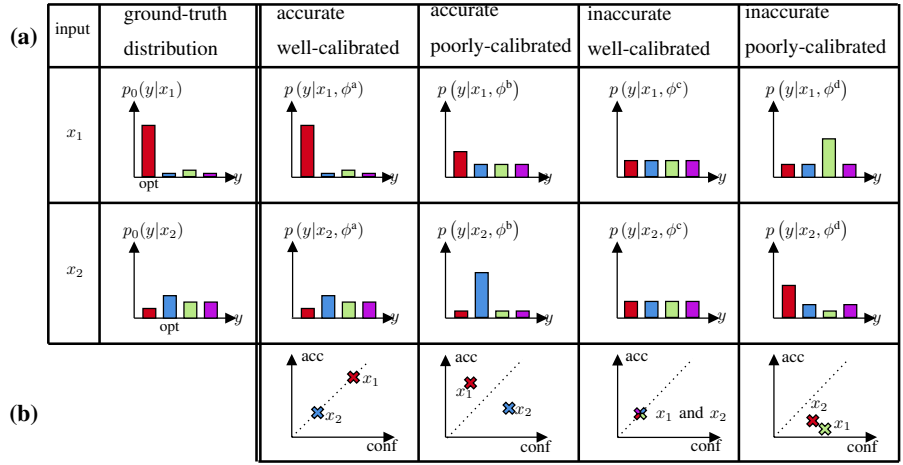

    \centering
    \includestandalone[trim=0cm 0cm 0cm 0cm, clip,width=12cm]{Figs/fig_tikz_4classes_acc_cal}
    \caption{\textbf{(a)} Examples of probabilistic predictors for two inputs $x_1$ and $x_2$: As compared to the ground-truth distribution in the second column, the first predictor (third column) is accurate, assigning the largest probability to the optimal decision (indicated as ``opt'' in the second column) and also well calibrated, reproducing the true accuracy of the decision; the second predictor (fourth column) is still accurate, but it is underconfident on the correct decision (for input $x_1$) and overconfident on the correct decision (for input $x_2$); the third predictor (fifth column) is not accurate, producing a uniform distribution across all output values, but is well calibrated if the data set is balanced \cite{vaicenavicius2019evaluating}; and the last predictor (sixth column) is both inaccurate and poorly calibrated, providing overconfident decisions. \textbf{(b)} Confidence versus accuracy for the decisions made by the corresponding predictors.}\label{fig: fig_tikz_4classes_acc_cal}
\end{figure*}

CP leverages probabilistic predictors to construct well-calibrated \emph{set predictors}. Instead of producing a probability vector, as in the examples in Fig.~\ref{fig: fig_tikz_4classes_acc_cal}, a set predictor outputs a subset of the output space, as exemplified in Fig.~\ref{fig: fig_tikz_4classes_covrg_ineff}. A set predictor is \emph{well calibrated} if it contains the correct output with a pre-defined \emph{coverage} probability selected by the system designer. For a well-calibrated set predictor, the size of the prediction set for a given input provides a measure of the uncertainty of the decision. Set predictors with smaller average prediction size are said to be more \emph{efficient} \cite{vovk2005algorithmic}.

This paper investigates CP as a general mechanism to obtain AI models with formal calibration guarantees for communication systems. The calibration guarantees of CP hold irrespective of the true, unknown, distribution underlying the generation of the variables of interest, and are defined either in terms of ensemble averages \cite{vovk2005algorithmic} or in terms of long-term averages \cite{gibbs2021adaptive}. CP is applied in conjunction to both frequentist and Bayesian learning, and specific applications are discussed to demodulation, modulation classification, and channel prediction.

\begin{figure}[t!]
    \centering
    \includestandalone[trim=0cm 0cm 0cm 0cm, clip,width=9cm]{Figs/fig_tikz_4classes_covrg_ineff}
    \caption{Set predictors produce subsets of the range of the output variable (here 

\tikzset{every picture/.style={line width=0.75pt}} 

\begin{tikzpicture}[x=0.75pt,y=0.75pt,yscale=-1,xscale=1]

\draw  [draw opacity=0][fill={rgb, 255:red, 217; green, 15; blue, 15 }  ,fill opacity=1 ][line width=0.75]  (19.7,4.68) -- (22.95,10.31) -- (26.2,15.93) -- (19.7,15.93) -- (13.2,15.93) -- (16.45,10.31) -- cycle ;
\draw  [draw opacity=0][fill={rgb, 255:red, 74; green, 144; blue, 226 }  ,fill opacity=1 ][line width=0.75]  (36.55,2.68) -- (38.42,6.93) -- (43.05,6.43) -- (40.3,10.18) -- (43.05,13.93) -- (38.42,13.43) -- (36.55,17.68) -- (34.67,13.43) -- (30.05,13.93) -- (32.8,10.18) -- (30.05,6.43) -- (34.67,6.93) -- cycle ;
\draw  [draw opacity=0][fill={rgb, 255:red, 126; green, 211; blue, 33 }  ,fill opacity=1 ][line width=0.75]  (53.4,2.68) -- (55.6,7.15) -- (60.53,7.86) -- (56.97,11.34) -- (57.81,16.25) -- (53.4,13.93) -- (48.99,16.25) -- (49.83,11.34) -- (46.27,7.86) -- (51.2,7.15) -- cycle ;
\draw  [draw opacity=0][fill={rgb, 255:red, 189; green, 16; blue, 224 }  ,fill opacity=1 ][line width=0.75]  (70.25,2.68) -- (72.9,7.53) -- (77.75,10.18) -- (72.9,12.83) -- (70.25,17.68) -- (67.6,12.83) -- (62.75,10.18) -- (67.6,7.53) -- cycle ;

\draw (45.49,10.18) node  [font=\normalsize] [align=left] {\{ \ \ \ \ \ \ \ \ \ \ \ \ \ \ \ \}};

\end{tikzpicture}

) for each input. Calibration is measured with respect to a desired coverage level $1-\alpha$: A set predictor is well calibrated if the true label is included in the prediction set with probability at least $1-\alpha$. A well-calibrated set predictor can be inefficient if it returns excessively large set predictions (forth column). In contrast, a poorly-calibrated set predictor (fifth column) returns set predictions that include the true value of the label with a probability smaller than $1-\alpha$.}\label{fig: fig_tikz_4classes_covrg_ineff}   
\end{figure} 

\subsection{Related Work}

Most work on AI for communications relies on conventional frequentist learning tools (see, e.g., the review papers \cite{simeone2018very,dai2020deep,shlezinger2021model,erpek2020deep}). \emph{Frequentist learning} is based on the minimization of the (regularized) training loss, which is interpreted as an estimate of the ground-truth population loss. When data is scarce, this estimate is unreliable, and hence the focus on a single, optimized, model parameter vector often yields probabilistic predictors that are poorly calibrated, producing overconfident decisions \cite{Guo2017Calibration,Sun2021amortized,cohen2022bayesian, zecchin2022robust}. 

\emph{Bayesian learning} offers a principled way to address this problem \cite{angelino2016patterns,simeone2022machine}. This is done by producing as the output of the learning process not a single model parameter vector, but rather a distribution in the model parameter space, which quantifies the model's epistemic uncertainty caused by limited access to data. A model trained via Bayesian learning produces probabilistic predictions that are averaged over the trained model parameter distribution. This \emph{ensembling} approach to prediction ensures that disagreements among models that fit the training data (almost) equally well are accounted for, substantially improving model calibration \cite{blundell2015weight,ravi2018amortized}.

In practice, Bayesian learning is implemented via approximations such as variational inference (VI) or Monte Carlo (MC) sampling, yielding scalable learning solutions \cite{simeone2022machine}. VI methods approximate the exact Bayesian posterior distribution with a tractable variational density \cite{graves2011practical,dusenberry2020efficient,daxberger2021laplace,farquhar2020liberty}, while MC techniques obtain approximate samples from the Bayesian posterior distribution \cite{neal2011mcmc,welling2011bayesian,zhang2019cyclical}. Among other applications to communications systems, Bayesian learning was studied for network allocation in \cite{narmanlioglu2017prediction,jha2021transformer,Maggi2021}, for massive MIMO detection in \cite{wu2022stochastic,zilberstein2022annealed,tao2021improved}, for channel estimation in \cite{jha2021online,prasad2015joint,lv2019joint}, for user identification in \cite{xu2021bayesian}, and for multi user detection in \cite{zhang2017bayesian,zhang2017novel}. Extensions to Bayesian meta-learning have been investigated in \cite{cohen2022bayesian}.

Exact Bayesian learning offers formal guarantees of calibration only under the assumption that the assumed model is \emph{well specified} \cite{morningstar2022pacm,masegosa2020learning}. In practice, this means that the assumed neural network models should have sufficient capacity to represent the ground-truth data generation mechanism, and that the predictive uncertainty should be unimodal for continuous outputs (since conventional likelihoods are unimodal, e.g., Gaussian) \cite{morningstar2022pacm, blundell2015weight, simeone2022machine}. These assumptions are easily violated in practice, especially in communication systems in which lower-complexity models must be implemented on edge devices, and access to data for specific network configurations is limited. Specific examples are provided in \cite{zecchin2022robust} for applications including modulation classification \cite{o2016convolutional, o2018over} and localization \cite{song2019novel, dvorecki2019machine}. 

Robustified versions of Bayesian learning that are based on the optimization of a modified free energy criterion were shown empirically to partly address the problem of model misspecification \cite{masegosa2020learning, morningstar2022pacm}, with implications for communication systems presented in \cite{zecchin2022robust}. However, robust Bayesian learning solutions do not have formal guarantees of calibration in the presence of misspecified models. 

Another family of methods that aim at enhancing the calibration of probabilistic models implement a validation-based post-processing phase. Platt scaling \cite{platt1999probabilistic} and temperature scaling \cite{Guo2017Calibration} find a fixed parametric mapping of the trained model output that minimizes the validation loss, while isotonic regression \cite{zadrozny2002transforming} applies a non-parametric binning approach. These recalibration-based approaches cannot guarantee calibration, as they may overfit the validation data set \cite{kumar2019verified} and they are sensitive to the inaccuracy of the starting model \cite{ma2021meta}.

\emph{Conformal prediction} is a general framework for the design of set predictors that satisfy formal, \emph{distribution-free}, guarantees of calibration \cite{vovk2005algorithmic, zeni2020conformal}. Given a desired miscoverage probability $\alpha$, CP returns set predictions that include the correct output value with probability at least $1-\alpha$ under the only assumption that the data distribution is \emph{exchangeable}. This condition is weaker that the standard assumption of ``i.i.d.'' data made in the design of most machine learning systems.

The original work on CP, \cite{vovk2005algorithmic}, introduced \emph{validation-based} CP and \emph{full} CP. Since then, progress has been made on reducing computational complexity, minimizing the size of the prediction sets, and further alleviating the assumptions of exchangeability. \emph{Cross-validation-based} CP was proposed in \cite{barber2021predictive} to reduce the computational complexity as compared to full CP, while improving the efficiency of validation-based CP. The authors of \cite{stutz2021learning, einbinder2022training} proposed the optimization of a \emph{CP-aware loss} to improve the efficiency of validation-based CP, while avoiding the larger computational cost of cross-validation. The work \cite{tibshirani2019conformal} proposed reweighting as a means to handle distribution shifts between the examples in the data set and the test point. Other research directions include improvements in the training algorithms \cite{yang2021finite,kumar2018training}, and the introduction of novel calibration metrics \cite{holland2020making,perez2022beyond}. Finally, \emph{online} CP, presented in \cite{gibbs2021adaptive, feldman2022conformalized}, was shown to achieve long-term calibration over time without requiring statistical assumptions on the data generation.

\subsection{Main Contributions}

To the best of our knowledge, with the exception of the conference version \cite{cohen2023calibrating} of this paper, this is the first work to investigate the application of CP to the design of AI models for communication systems. 
The main contributions of this paper are as follows.
\begin{itemize}
\item We provide a self-contained introduction to CP by focusing on validation-based CP \cite{vovk2005algorithmic}, cross-validation-based CP \cite{barber2021predictive}, and online conformal prediction \cite{feldman2022conformalized}. The presentation details connections to conventional probabilistic predictors, as well as the  performance metrics used to assess calibration and efficiency.
\item We propose the application of offline CP to the problems of symbol demodulation and modulation classification. The experimental results validate the theoretical property of CP methods of providing well-calibrated decisions. Furthermore, they demonstrate that \naive predictors that only rely on the output of either frequentist or Bayesian learning tools often result in poor calibration.
\item Finally, we study the application of online CP to the problem of predicting received signal strength for over-the-air measured signals \cite{simmons2022ai}. We demonstrate that online CP can obtain the predefined target long-term coverage rate at the cost of negligible increase in the prediction interval as compared to \naive predictors.
\end{itemize}
The conference version \cite{cohen2023calibrating} of this work presented results only for symbol demodulation, while not providing background material on CP and not considering online CP. In contrast, this work is self-contained, presenting CP from first principles and including also online CP. Furthermore, this work investigates applications of CP to modulation classification and to channel prediction by leveraging real-world data sets \cite{zanella2013rss,simmons2022ai}. For reproducibility purposes, we have made our code publicly available\footnote{\url{https://github.com/kclip/cp4wireless}}.

The rest of this paper is organized as follows. In Sec.~\ref{sec: Problem Definition}, we define set predictors, and introduce the relevant performance metrics. Then, in Sec.~\ref{sec: Set Predictors}, \naive set predictors are introduced that do not provide  guarantees in terms of  calibration. Sec.~\ref{sec: Conformal Prediction} describes conformal prediction, a general methodology to obtain well-calibrated set predictors. Sec.~\ref{sec: Online Conformal Prediction} details online conformal prediction, which is well suited for time-varying data. Applications to wireless communications are investigated in the following sections: Symbol demodulation is studied in Sec.~\ref{sec: Symbol Demodulation}; modulation classification in Sec.~\ref{sec: Modulation Classification}; and channel prediction in Sec.~\ref{sec: Online Channel Prediction}. Sec.~\ref{sec: conclusions} concludes the paper.

\section{Problem Definition}\label{sec: Problem Definition}

This section introduces set predictors, along with key performance metrics of coverage and inefficiency. To this end, we start by describing the data-generation model and reviewing probabilistic predictors.

\subsection{Data-Generation Model}

We consider the standard supervised learning setting in which the learner is given a data set $\D=\{ z[i] \}_{i=1}^N$ of $N$ examples of input-output pairs $z[i]=(x[i],y[i])$ for $i=1,\ldots,N$, and is tasked with producing a prediction on a test input $x$ with unknown output $y$. Writing $z=(x,y)$ for the test pair, data set $\D$ and test point $z$ follow the unknown \emph{ground-truth}, or \emph{population}, \emph{distribution} $p_0(\D,z)$. Apart from Sec.~\ref{sec: Online Conformal Prediction}, we further assume throughout that the population distribution $p_0(\D,z)$ is \emph{exchangeable} -- a condition that includes as a special case the traditional independent and identically distributed (i.i.d.) data-generation setting. Note that we will not make explicit the distinction between random variables and their realizations, which will be clear from the context.

Mathematically, exchangeability requires that the joint distribution $p_0(\D,z)$ does not depend on the ordering of the $N+1$ variables $\{z[1],\dots,z[N],z\}$. Equivalently, by de Finetti's theorem \cite{definneti1974theory}, there exists a latent random vector $\rv{c}$ with distribution $p_0(c)$ such that, conditioned on $\rv{c}$, the variables $\{z[1],\dots,z[N],z\}$ are i.i.d. Writing the conditional i.i.d. distribution as 
\begin{equation}
    p_0(\D,z|\rv{c})  = p_0(z|\rv{c}) \prod_{i=1}^N  p_0\big(z[i]\big|\rv{c}\big)
\end{equation}
for some ground-truth sampling distribution $p_0(z|c)$ given the variable $c$, under the exchangeability assumption, the joint distribution can be expressed as 
\begin{align}
    p_0(\D,z) = \E_{p_0(c)} \Big[ p_0(\D,z|\rv{c})  \Big],\label{eq: exchangeability} 
\end{align}
where $\E_{p(x)}[\cdot]$ denotes the expectation with respect to distribution $p(x)$.

The vector $\rv{c}$ in \eqref{eq: exchangeability} can be interpreted as including context variables that determine the specific learning task.  For instance, in a wireless communication setting, the vector $\rv{c}$ may encode information about channel conditions. In Sec.~\ref{sec: Online Conformal Prediction}, we will consider a more general setting in which no assumptions are made on the distribution of the data.

\subsection{Probabilistic Predictors}\label{sec: Probabilistic Predictors}

Before introducing set predictors, we briefly review conventional probabilistic predictors. \emph{Probabilistic predictors} implement a parametric conditional distribution model $p(y|x,\phi)$ on the output $y \in \mathcal{Y}$ given the input $x \in \mathcal{X}$, where $\phi\in\Phi$ is a vector of model parameters. Given the training data set $\D$, frequentist learning produces an optimized single vector $\phi_\D^*$, while Bayesian learning returns a distribution $q^*(\phi|\D)$ on the model parameter space $\Phi$ \cite{blundell2015weight,simeone2022machine}. In either case, we will denote as $p(y|x,\D)$ the resulting optimized predictive distribution
\begin{equation}
    p(y|x,\D) = 
    \begin{cases}
        p(y|x,\phi_{\D}^*)                                   & \text{for frequentist learning} \\
        \E_{q^*(\phi|\D)} [p(y|x,\rvphi)]    & \text{for Bayesian learning} .
    \end{cases} \label{eq: p(y|x,D)}
\end{equation}
Note that the predictive distribution for Bayesian learning is obtained by averaging, or ensembling, over the optimized distribution $q^*(\phi|\D)$. We refer to Appendix~\ref{appendix: Frequentist and Bayesian Learning} for basic background on frequentist and Bayesian learning.

From \eqref{eq: p(y|x,D)}, one can obtain a point prediction $\hat{y}$ for output $y$ given input $x$ as the probability-maximizing output as
\begin{equation}
    \hat{y}(x|\D) = \argmax_{y^\prime\in\mathcal{Y}} p(y^\prime|x,\D). \label{eq:hard pred}
\end{equation}
In the case of a discrete set $\mathcal{Y}$, the hard predictor \eqref{eq:hard pred} minimizes the probability of detection error under the model $p(y|x,\D)$. The probabilistic prediction $p(y|x,\D)$ also provides a measure of predictive uncertainty for all possible outputs $y\in \mathcal{Y}$. In particular, for the point prediction $\hat{y}(x|\D)$ in \eqref{eq:hard pred}, we have the predictive, self-reported, \emph{confidence level}
\begin{equation}
    \mathrm{conf}(x|\D) = \max_{y^\prime\in\mathcal{Y}} p(y^\prime|x,\D) = p\big(\hat{y}(x|\D)\big|x,\D\big) . \label{eq:pred_conf}
\end{equation}

As illustrated in Fig.~\ref{fig: fig_tikz_4classes_acc_cal}, the performance of a probabilistic predictor can be evaluated in terms of both accuracy and \emph{calibration}, with the latter quantifying the quality of uncertainty quantification via the confidence level \eqref{eq:pred_conf} \cite{Guo2017Calibration}. Specifically, a probabilistic predictor $p(y|x,\D)$ is said to be \emph{well calibrated} \cite{Guo2017Calibration} if the probability that the hard predictor $\hat{y}=\hat{y}(x|\D)$ equals the true label matches its confidence level $\pi$ for all possible values of probability $\pi \in [0,1]$. Mathematically, calibration is defined by the condition 
\begin{equation}
    \Prob\big( \rv{y} = \hat{y} \big| p(\hat{y}|\rv{x},\D) = \pi \big) = \pi, \quad \text{ for all }\pi\in[0,1]
    \label{eq:wc_for_prob}
\end{equation}
where the probability $\Prob(\cdot)$ follows the ground-truth distribution $p_0(x,y)$. Stronger definitions, like that introduced in \cite{park2020calibrated}, require the predictive distribution to match the ground-truth distribution also for values of $y$ that are distinct from \eqref{eq:hard pred}.

\subsection{Set Predictors}

A \emph{set predictor} is defined as a set-valued function $\Gamma(\cdot|\D): ~\mathcal{X} ~\rightarrow ~2^\mathcal{Y}$ that maps an input $x$ to a subset of the output domain $\mathcal{Y}$ based on data set $\D$. We denote the size of the set predictor for input $x$ as $|\Gamma(x|\D)|$. As illustrated in the example of Fig.~\ref{fig: fig_tikz_4classes_covrg_ineff}, the set size $|\Gamma(x|\D)|$ generally depends on input $x$, and it can be taken as a measure of the uncertainty of the set predictor.

The performance of a set predictor is evaluated in terms of calibration, or coverage, as well as of inefficiency. \emph{Coverage} refers to the probability that the true label is included in the predicted set; while \emph{inefficiency} refers to the average size $|\Gamma(x|\D)|$ of the predicted set. There is clearly a trade-off between two metrics. A conservative set predictor that always produces the entire output space, i.e., $\Gamma(x|\D)=\mathcal{Y}$, would trivially yield a coverage probability equal to $1$, but at the cost of exhibiting the worst possible inefficiency of $|\mathcal{Y}|$. Conversely, a set predictor that always produces an empty set, i.e., $\Gamma(x|\D)=\emptyset$, would achieve the best possible inefficiency, equal to zero, while also presenting the worst possible coverage probability equal to zero. 

Let us denote a set predictor $\Gamma(\cdot|\cdot)$ for short as $\Gamma$. Formally, the \emph{coverage} level of set predictor $\Gamma$ is the probability that the true output $y$ is included in the prediction set $\Gamma(x|\D)$ for a test pair $z=(x,y)$. This can be expressed as $\mathrm{coverage}(\Gamma) ~= \Prob~\big(\rv{y}\in \Gamma(\rv{x}|\rvD)\big)$, where the probability $\Prob(\cdot)$ is taken over the ground-truth joint distribution $p_0(\D,(x,y))$ in \eqref{eq: exchangeability}. The set predictor $\Gamma$ is said to be $(1-\alpha)$-\emph{valid} if it satisfies the inequality
\begin{equation}
    \mathrm{coverage}(\Gamma) = \Prob\big(\rv{y} \in \Gamma(\rv{x}|\rvD)\big) \geq 1-\alpha.
    \label{eq: set validity}
\end{equation}
When the desired coverage level $1-\alpha$ is fixed by the predetermined target \emph{miscoverage level} $\alpha \in [0,1]$, we will also refer to set predictors satisfying \eqref{eq: set validity} as being well calibrated. 

Following the discussion in the previous paragraph, it is straightforward to design a valid, or well-calibrated, set predictor, even for the restrictive case of miscoverage level $\alpha=0$. This can be, in fact, achieved by producing the full set $\Gamma(x|\D)=\mathcal{Y}$ for all inputs $x$. One should, therefore, also consider the inefficiency of predictor $\Gamma$. The \emph{inefficiency} of set predictor $\Gamma$ is defined as the average prediction set size
\begin{equation}
    \mathrm{inefficiency}(\Gamma) = \E \Big[ \big|\Gamma(\rv{x}|\rvD)\big| \Big], \label{eq: ineff(Gamma) = E | Gamma |}
\end{equation}
where the average is taken over the data set $\rvD$ and the test pair $(\rv{x},\rv{y})$ following their exchangeable joint distribution $p_0(\D,(x,y))$.

\begin{figure}
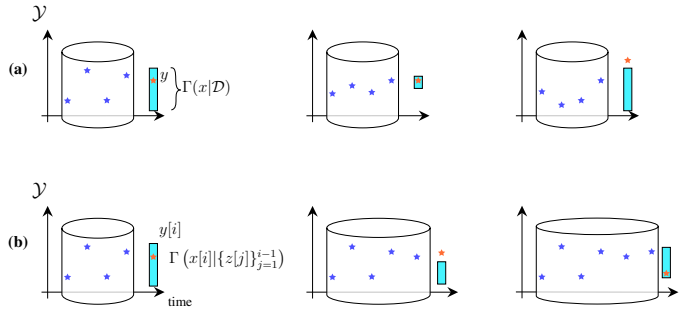

    \centering
    \includestandalone[width=9cm]{Figs/fig_tikz_cp_offline_online}
    \caption{\textbf{(a)} The validity condition \eqref{eq: set validity} assumed in offline CP is relevant if one is interested in the average performance with respect to realizations $(\rvD,\rv{z})~\sim~ p_0(\D,z)$ of training set $\D$ and test variable $z=(x,y)$. Input variable $x$ is not explicitly shown in the figure, and the horizontal axis runs over the training examples in $\D$ and the test example $z$. \textbf{(b)} In online CP, the set predictor $\Gamma_i$ uses its input $x[i]$ and all previously observed pairs $z[1],\dots,z[i-1]$ with $z[i]=(x[i],y[i])$ to produce a prediction set. The long-term validity \eqref{eq: long-term set validity} assumed by online CP is defined as the empirical time-average rate at which the predictor $\Gamma_i$ includes the true target variable $y[i]$.}
    \label{fig: fig_tikz_cp_offline_online}
\end{figure}

In practice, the coverage condition \eqref{eq: set validity} is relevant if the learner produces multiple predictions using independent data set $\D$, and is tested on multiple pairs $(x,y)$. In fact, in this case, the probability in \eqref{eq: set validity} can be interpreted as the fraction of predictions for which the set predictor $\Gamma(x|\D)$ includes the correct output. This situation, illustrated in Fig.~\ref{fig: fig_tikz_cp_offline_online}(a), is quite common in communication systems, particularly at the lower layers of the protocol stack. For instance, the data $\D$ may correspond to pilots received in a frame, and the test point $z$ to a symbol within the payload part of the frame (see Sec.~\ref{sec: Symbol Demodulation}). While the coverage condition \eqref{eq: set validity} is defined under the assumption of a fixed ground-truth distribution $p_0(\D,z)$, in Sec.~\ref{sec: Online Conformal Prediction} we will allow for temporal distributional shifts and we will focus on validity metrics defined as long-term time averages (see Fig.~\ref{fig: fig_tikz_cp_offline_online}(b)).

\section{\Naive Set Predictors}\label{sec: Set Predictors}

Before describing CP in the next section, in this section we review two \naive, but natural and commonly used, approaches to produce set predictors, that fail to satisfy the coverage condition \eqref{eq: set validity}.

\subsection{\Naive Set Predictors from Probabilistic Predictors}\label{sec: Set Predictors from Probabilistic Predictors}

\begin{figure}
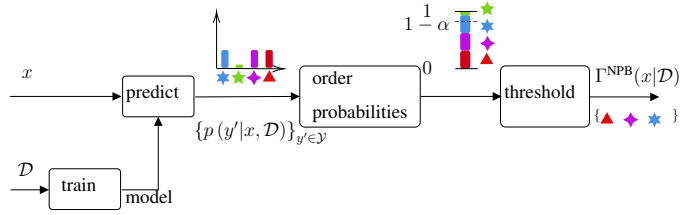

    \centering
    \includestandalone[width=9cm]{Figs/fig_tikz_NPB_setpred_blockdiagram}
    \caption{A \naive probabilistic-based (NPB) set predictor uses a pre-trained probabilistic predictor to include all output values to which the probabilistic predictor assigns the largest probabilities that reach the coverage target $1-\alpha$. This \naive scheme has no formal guarantee of calibration, i.e., it does not guarantee the coverage condition \eqref{eq: set validity}, unless the original probabilistic predictor is well calibrated.}
    \label{fig: fig_tikz_NPB_setpred_blockdiagram}
\end{figure}

Given a probabilistic predictor $p(y|x,\D)$ as in \eqref{eq: p(y|x,D)}, one could construct a set predictor by relying on the confidence levels reported by the model. Specifically, aiming at satisfying the coverage condition \eqref{eq: set validity}, given an input $x$, one could construct the smallest subset of the output domain $\mathcal{Y}$ that covers a fraction $1-\alpha$ of the probability designed by model $p(y|x,\D)$. Mathematically, the resulting \emph{\naive probabilistic-based} (NPB) set predictor is defined as
\begin{IEEEeqnarray}{rcl}
        \Gamma^{\text{NPB}}(x|\D) = &\underset{\Gamma \in 2^\mathcal{Y}}{\argmin}\: & |\Gamma| \label{eq: prediction set naive} \\
        &\text{s.t.} & \sum_{y^\prime \in \Gamma} p(y^\prime|x,\D) \geq 1-\alpha \nonumber
\end{IEEEeqnarray}
for the case of a discrete set, and an analogous definition applies in the case of a continuous domain $\mathcal{Y}$. Fig.~\ref{fig: fig_tikz_NPB_setpred_blockdiagram} illustrates the NPB for a  prediction problem with output domain size $|\mathcal{Y}|=4$.
Given that, as mentioned in Sec.~\ref{sec: Introduction}, probabilistic predictors are typically poorly calibrated, the \naive set predictor \eqref{eq: prediction set naive} does not satisfy condition \eqref{eq: set validity} for the given desired miscoverage level $\alpha$, and hence it is not well calibrated. For example, in the typical case in which the probabilistic predictor is overconfident \cite{Guo2017Calibration}, the predicted sets \eqref{eq: prediction set naive} tend to be too small to satisfy the coverage condition \eqref{eq: set validity}.

\subsection{\Naive Set Predictors from Quantile Predictors}\label{sec: Set Predictors from Quantile Predictors}

While the \naive probabilistic-based set predictor \eqref{eq: prediction set naive} applies to both discrete and continuous target variables, we now focus on the important special case in which $\mathcal{Y}$ is a real number, i.e., $\mathcal{Y}=\mathbb{R}$. This corresponds to scalar regression problems, such as for channel prediction (see Sec.~\ref{sec: Online Channel Prediction}). Under this assumption, one can construct a \naive set predictor based on estimates of the $\alpha/2$- and $(1-\alpha/2)$-quantiles $y_{\alpha/2}(x)$ and $y_{1-\alpha/2}(x)$ of the ground-truth distribution $p_0(y|x)$ (obtained from the joint distribution $p_0(\D,z)$). In fact, writing as
\begin{equation}
    y_q(x) = \inf \Big\{ y\in\mathbb{R} : \int_{-\infty}^y p_0(y^\prime|x) \dd y^\prime \leq q \big\} \label{eq: y_q(x)}
\end{equation}
the $q$-quantile, with $q\in[0,1]$, of the ground-truth distribution $p_0(y|x)$, the interval $\big[ y_{\alpha/2}(x) , y_{1-\alpha/2}(x) \big]$ contains the true value $y$ with probability $1-\alpha$.

Defining the pinball loss as \cite{koenker1978regression} 
\begin{equation}
    \ell_q(y,\hat{y})=\max\big\{-(1-q)(y-\hat{y}), q(y-\hat{y})\big\} \label{eq: def pinball loss}
\end{equation}
for $q\in[0,1]$, the quantile $y_q(x)$ in \eqref{eq: y_q(x)} can be obtained as \cite{steinwart2011estimating}
\begin{equation}
    y_q(x) = \underset{\hat{y}\in\mathbb{R}}{\argmin}\: \E_{p_0(y|x)} \big[ \ell_q(\rv{y},\hat{y}) \big].
\end{equation}
Therefore, given a parametrized predictive model $\hat{y}(x|\phi)$, the quantile $y_q(x)$ can be estimated as $\hat{y}(x|\phi_{\D,q})$ with optimized parameter vector
\begin{equation}
    \phi_{\D,q} = \argmin_\phi \bigg\{ \tfrac{1}{N} \sum_{(x,y)\in\D} \ell_q \big(y,\hat{y}(x|\phi) \big) \bigg\}.
    \label{eq: phi_lo and phi_hi as pinball minimizers}
\end{equation}
With the estimate $\hat{y}(x|\phi_{\D,\alpha/2})$ of quantile $y_{\alpha/2}(x)$ and estimate $\hat{y}(x|\phi_{\D,1-\alpha/2})$ of quantile $y_{1-\alpha/2}(x)$, we finally obtain the \emph{\naive quantile-based} (NQB) predictor
\begin{equation}
    \Gamma^\text{NQB} (x|\D) = \Big[ \hat{y}(x|\phi_{\D,\alpha/2}) , \hat{y}(x|\phi_{\D,1-\alpha/2}) \Big] . \label{eq: naive quantile based}
\end{equation} 
The \naive set prediction in \eqref{eq: naive quantile based} fails to satisfy the condition \eqref{eq: set validity}, since the empirical quantiles $\hat{y}_q(x)$ generally differ from the ground-truth quantiles $y_q(x)$.

\section{Conformal Prediction}\label{sec: Conformal Prediction}

In this section, we review CP-based set predictors, which have the key property of guaranteeing the $(1-\alpha)$-validity condition \eqref{eq: set validity} for any predetermined miscoverage level $\alpha$, irrespective of the ground-truth distribution $p_0(\D,z)$ of the data. We specifically focus on validation-based CP \cite{vovk2005algorithmic} and cross-validation-based CP \cite{barber2021predictive}, which are more practical variants of full CP \cite{vovk2005algorithmic,lei2018distribution}. In Sec.~\ref{sec: Online Conformal Prediction}, we cover online CP \cite{gibbs2021adaptive, feldman2022conformalized}.

\subsection{Validation-Based CP (VB-CP)}\label{sec: Validation-Based CP}

\begin{figure}
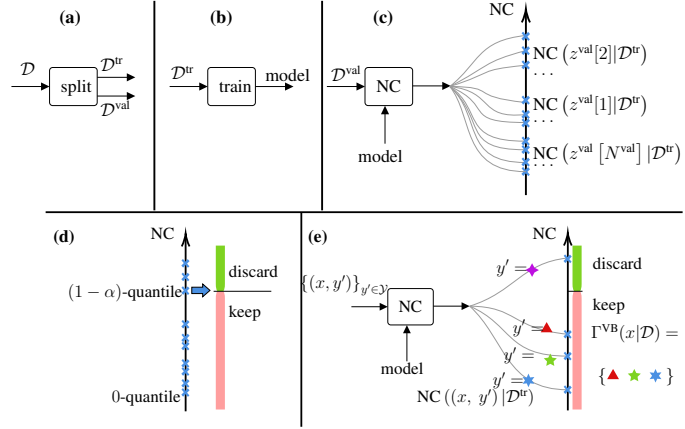

    \centering
    \includestandalone[width=9cm]{Figs/fig_tikz_VB_CP_comic_LHS}
    \vspace{0.5cm}
    \includestandalone[width=9cm]{Figs/fig_tikz_VB_CP_comic_RHS}
    \caption{Validation-based conformal prediction (VB-CP): \textbf{(a)} The data set is split into training and validation set; \textbf{(b)} A single model is trained over the training data set; \textbf{(c)-(d)} Post-hoc calibration is done by evaluating the NC scores on the validation set \textbf{(c)} and by identifying the $(1-\alpha)$-quantile of the validation NC scores. This divides the axis of NC scores into a ``keep'' region of NC scores smaller than the threshold, and into a complementary ``discard'' region \textbf{(d)}. \textbf{(e)} For each test input $x$, VB-CP includes in the prediction set all labels $y^\prime\in\mathcal{Y}$ for which the NC score of the pair $(x,y^\prime)$ is within the ``keep'' region.}
    \label{fig: fig_tikz_VB_CP_comic}
\end{figure} 

In this subsection, we describe \emph{validation-based} CP (VB-CP), which partitions the available set $\D=\Dtr\cup\Dval$ into a training set $\Dtr$ with $\Ntr$ samples and a validation set $\Dval$ with $\Nval=N-\Ntr$ samples (Fig.~\ref{fig: fig_tikz_VB_CP_comic}(a)). This class of methods is also known as inductive CP \cite{vovk2005algorithmic} or split CP \cite{barber2021predictive}.

VB-CP operates on any pre-trained probabilistic model $p(y|x,\Dtr)$ obtained using the training set $\Dtr$ as per \eqref{eq: p(y|x,D)}. At test time, given an input $x$, VB-CP relies on a validation set to determine which labels $y^\prime \in \mathcal{Y}$ should be included in the predicted set. Specifically, for any given test input $x$, a label $y^\prime\in\mathcal{Y}$ is included in set $\Gamma^\text{VB}(x|\D)$ depending on the extent to which the candidate pair $(x,y^\prime)$ ``conforms'' with the examples in the validation set. 

This ``conformity'' test for a candidate pair is based on a \emph{nonconformity (NC) score}. An NC score for VB-CP can be obtained as the log-loss
\begin{equation}
    \NC(z=(x,y)|\Dtr) = -\log p(y|x,\Dtr) \label{eq: NC classification}
\end{equation}
or as any other score function that measures the loss of the probabilistic predictor $p(y|x,\Dtr)$ on example $(x,y)$.  It is also possible to define NC scores for quantile-based predictors as in \eqref{eq: naive quantile based}, and we refer to \cite{feldman2022conformalized} for details.

VB-CP consists of a training phase (Fig.~\ref{fig: fig_tikz_VB_CP_comic}(a)-(d)) and of a test phase (Fig.~\ref{fig: fig_tikz_VB_CP_comic}(e)). During \emph{training}, the data set $\Dtr$ is used to obtain a probabilistic predictor $p(y|x,\Dtr)$ as in \eqref{eq: p(y|x,D)} (Fig.~\ref{fig: fig_tikz_VB_CP_comic}(b)). Then, NC scores $\NC(z^\text{val}[i]|\Dtr)$, as in \eqref{eq: NC classification}, are evaluated on all points $z^\text{val}[i], i=1,\ldots,\Nval$ in the validation set $\Dval$ (Fig.~\ref{fig: fig_tikz_VB_CP_comic}(c)). Finally, the real line of NC scores is partitioned into a ``keep'' region and a ``discard'' region (Fig.~\ref{fig: fig_tikz_VB_CP_comic}(d)), choosing as a threshold the $(1-\alpha)$-empirical quantile of the $\Nval$ NC scores $\{\NC(z^\text{val}[i]|\Dtr)\}_{i=1}^{\Nval}$. Accordingly, we ``keep'' the labels $y^\prime$ with NC scores that are smaller than the $(1-\alpha)$-empirical quantile of the validation NC scores, and ``discard'' larger NC scores.

During \emph{testing} (Fig.~\ref{fig: fig_tikz_VB_CP_comic}(e)), given a test input $x$, $|\mathcal{Y}|$ NC scores are evaluated, one for each of the candidate labels $y^\prime\in\mathcal{Y}$, using the same trained model $p(y|x,\Dtr)$. All candidate labels $y^\prime$ for which the NC score $\NC((x,y^\prime)|\Dtr)$ falls within the ``keep'' region are included in the predicted set of VB-CP.

Mathematically, the VB-CP set predictor is obtained as
\begin{IEEEeqnarray}{rcl}
    \Gamma^\text{VB}(x|\D) =  \Big\{ y^\prime\in\mathcal{Y} \Big| && \NC((x,y^\prime)|\Dtr)  \label{eq: prediction set VB}\\ 
    \negspaceF&& \leq \quan_\alpha \big(\{ \NC(z^\text{val}[i]|\Dtr)\}_{i=1}^{N^\text{val}}\big) \Big\}, \nonumber
\end{IEEEeqnarray}
where the \emph{empirical quantile from the top} for a set of $N$ real values $\{r[i]\}_{i=1}^{N}$ is defined as
\begin{IEEEeqnarray}{rcl}
    \quan_\alpha\ \big(\{r[i]\}_{i=1}^{N}\big)&=& \big\lceil  (1-\alpha)(N+1) \big\rceil \text{th smallest value} \nonumber \\
    && \text{         of the set } \{r[i]\}_{i=1}^{N}\cup\{+\infty\}. \label{eq: quantiles of a vector}
\end{IEEEeqnarray}

\subsection{Cross-Validation-Based CP (CV-CP)}\label{sec: Cross-Validation-Based Set Predictors}

\begin{figure*}
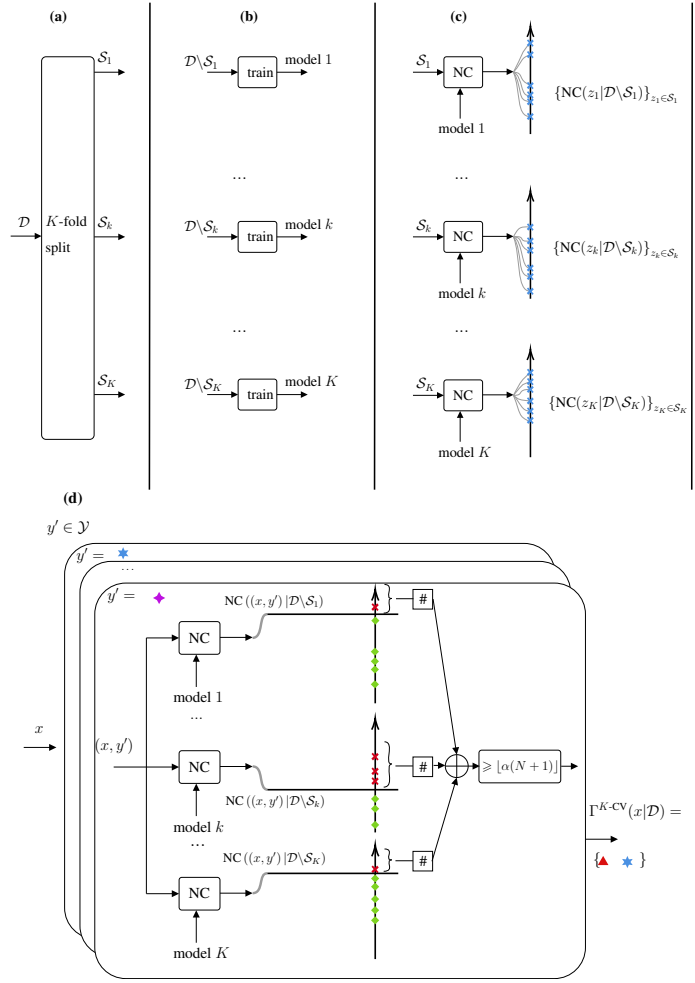

    \centering
    \includestandalone[height=6.5cm]{Figs/fig_tikz_KCV_CP_comic_LHS}
    \includestandalone[height=6.5cm]{Figs/fig_tikz_KCV_CP_comic_RHS}
    \caption{$K$-fold cross-validation-based conformal prediction ($K$-CV-CP): \textbf{(a)} The $N$ data pairs of data set $\D$ are split into $K$-folds each with $|\mathcal{S}_k|=N/K$ samples; \textbf{(b)} $K$ models are trained, each using a leave-fold-out data set of $|\D\setminus\mathcal{S}_k|=N-N/K$ pairs; \textbf{(c)} NC scores are computed on the $N/K$ holdout data points for each fold $\mathcal{S}_k$; \textbf{(d)} For each test input $x$, all labels $y^\prime\in\mathcal{Y}$ for which the number of ``higher-NC'' validation points exceeds a fraction $\alpha$ of the total $N$ points are considered in the prediction set. CV-CP is the special case with $K=N$.}
    \label{fig: fig_tikz_KCV_CP_comic}
\end{figure*}

VB-CP has the computational advantage of requiring the training of a single model, but the split into training and validation data causes the available data to be used in an inefficient way. This data inefficiency generally yields set predictors with a large average size \eqref{eq: ineff(Gamma) = E | Gamma |}. Unlike VB-CP, \emph{cross-validation-based} CP (CV-CP) \cite{barber2021predictive} trains multiple models, each using a subset of the available data set $\D$. As detailed next and summarized in Fig.~\ref{fig: fig_tikz_KCV_CP_comic}, during the training phase, each data point $z[i]$ in the validation set is assigned an NC score based on a model trained using a subset of the data set $\D$ that excludes $z[i]$, with $i\in \{1,...,N\}$. Then, for testing, the inclusion of a label $y^\prime$ in the prediction set for an input $x$ is based on a comparison of NC scores evaluated for the pair $(x,y^\prime)$ with all the $N$ validation NC scores. 

Specifically, as illustrated in Fig.~\ref{fig: fig_tikz_KCV_CP_comic}, $K$-fold CV-CP \cite{barber2021predictive}, referred here as $K$-CV-CP, first partitions the data set $\D$ into $K$ disjoint folds $\{ \mathcal{S}_k\}_{k=1}^K$, each with $N/K$ points, i.e., $\cup_{k=1}^K \mathcal{S}_k~=~\D$ (Fig.~\ref{fig: fig_tikz_KCV_CP_comic}(a)), for a predefined integer $K\in\{2,\dots,N\}$ such that the ratio $N/K$ is an integer. 

During \emph{training}, the $K$ subsets $\D\setminus\mathcal{S}_k$ are used to train $K$ probabilistic predictors $p(y|x,\D\setminus\mathcal{S}_{k})$ defined as in \eqref{eq: p(y|x,D)} (Fig.~\ref{fig: fig_tikz_KCV_CP_comic}(b)). Each trained model $p(y|x,\D\setminus\mathcal{S}_{k})$ is used to evaluate the $|\mathcal{S}_k|=N/K$ NC scores $\NC\big(z_k\big|\D\setminus\mathcal{S}_{k}\big)$ for all validation data points $z_k\in\mathcal{S}_k$ that were not used for training the model (Fig.~\ref{fig: fig_tikz_KCV_CP_comic}(c)). Unlike VB-CP, $K$-CV-CP requires  keeping in memory all the $N$ validation scores for testing. These points are illustrated as crosses in Fig.~\ref{fig: fig_tikz_KCV_CP_comic}(c).

During \emph{testing}, for a given test input $x$  and for any candidate label $y^\prime\in\mathcal{Y}$, CV-CP evaluates $K$ NC scores, one for each of the $K$ trained models. Each such NC score $\NC\big((x,y^\prime)\big|\D\setminus\mathcal{S}_{k}\big)$ is compared with the $N/K$ validation scores obtained on fold  $\mathcal{S}_{k}$. We then count how many of the $N/K$  validation scores are larger than $\NC\big((x,y^\prime)\big|\D\setminus\mathcal{S}_{k}\big)$. If the sum of all such counts, across the $K$ folds  $\{ \mathcal{S}_k\}_{k=1}^K$, is larger than a fraction $\alpha$ of all $N$ data points, then the candidate label $y^\prime$ is included in the prediction set (Fig.~\ref{fig: fig_tikz_KCV_CP_comic}(d)). This criterion follows the same principle of VB-CP of including all candidate labels $y^\prime$ that ``conform'' well with a sufficiently large fraction of validation points.

Mathematically, $K$-CV-CP is defined as
\begin{IEEEeqnarray}{rcl}
    \Gamma^\text{$K$-CV}(x|\D) =  \bigg\{ y^\prime\in\mathcal{Y} \bigg| && \sum_{k=1}^K \sum_{z_k\in\mathcal{S}_k} \indicator \Big( \NC\big((x,y^\prime)\big|\D\setminus\mathcal{S}_k\}\big) \label{eq: prediction set KCV-CP classification} \\
    && \leq \NC\big(z_k\big|\D\setminus\mathcal{S}_k\big) \Big) \geq \left\lfloor\alpha (N+1)\right\rfloor \bigg\}, \nonumber
\end{IEEEeqnarray}
where $\indicator(\cdot)$ is the indicator function ($\indicator(\text{true})=1$ and $\indicator(\text{false})=0$). The left-hand side of the inequality in \eqref{eq: prediction set KCV-CP classification} implements the sums, shown in Fig.~\ref{fig: fig_tikz_KCV_CP_comic}(d), over counts of validation NC scores that are larger than the corresponding NC score for the candidate pair $(x,y^\prime)$. 

$K$-CV-CP increases the computational complexity $K$-fold as compared to VB-CP, while generally reducing the inefficiency \cite{barber2021predictive}. The special case 
 of $K=N$, known as jackknife+ \cite{barber2021predictive}, is referred here as CV-CP. In this case, each of the $N$ folds $\mathcal{S}_k,k=1,\dots,N$  uses a single cross validation point. In general, CV-CP is the most efficient form of $K$-CV-CP, but it may be impractical for large data set sizes due to need to train $N$ models. The number of folds $K$ should strike a balance between computational complexity, as $K$ models are trained,  and inefficiency.

\subsection{Calibration Guarantees}\label{sec: Calibration Guarantees}

 VB-CP \eqref{eq: prediction set VB} satisfies the coverage condition \eqref{eq: set validity} \cite{vovk2005algorithmic} under the only assumption of exchangeability (see Sec. II-A).

The validity of CV-CP requires a technical assumption on the NC score. While in VB-CP the NC score is an arbitrary score function evaluated based on any pre-trained probabilistic model, for CV-CP, the NC score $\NC(z|\D^\prime)$ must satisfy the additional property of being invariant to permutations of the data set $\D^\prime$ used to train the underlying probabilistic model. Consider the log-loss $\NC(z=(x,y)|\D^\prime) = -\log p(y|x,\D^\prime)$ \eqref{eq: NC classification}, or any other score function based on the trained model $p(y|x,\D^\prime)$, as the NC score. CV-CP requires that  the training algorithm used to produce model $p(y|x,\D^\prime)$ provides outputs that are invariant to permutations of the training set $\D^\prime$. 

Specifically, for frequentist learning, the optimization algorithm producing the parameter vector $\phi_{\D^\prime}^*$ in \eqref{eq: p(y|x,D)} must be permutation-invariant. This is the case for standard methods such as  full-batch gradient descent (GD), or for non-parametric techniques such as Gaussian processes. For Bayesian learning, the distribution $q^*(\phi|\D^\prime)$ in \eqref{eq: p(y|x,D)} must also be permutation-invariant, which is true for the exact posterior distribution \cite{simeone2022machine}, as well as for approximations obtained via MC methods such as Langevin MC \cite{welling2011bayesian,simeone2022machine}.

The requirement on permutation-invariance can be alleviated by allowing for probabilistic training algorithms such as stochastic gradient descent (SGD) \cite{barber2022conformal}. With probabilistic training algorithms, the only requirement is that the \emph{distribution} of the (random) output models is permutation-invariant. This is, for instance, the case if SGD is implemented by taking mini-batches uniformly at random within the training set $\D^\prime$ \cite{barber2022conformal, romano2020classification, wang2022probabilistic}. With probabilistic training algorithms, however, the validity condition \eqref{eq: set validity} of CV-CP is only guaranteed on average with respect to the random outputs of the algorithms.

Specifically, under the discussed assumption of permutation-invariance of the NC scores, by \cite[Theorems 1 and 4]{barber2021predictive}, CV-CP satisfies the inequality
\begin{equation}
    \Prob\big( \rv{y} \in \Gamma^\text{CV}(\rv{x}|\rvD) \big) \geq 1-2\alpha, \label{eq: CV-CP is valid}
\end{equation}
while $K$-CV-CP satisfies the inequality
\begin{IEEEeqnarray}{rcl}
    \Prob\big( \rv{y} \in \Gamma^\text{$K$-CV}(\rv{x}|\rvD) \big) 
    &\geq& 1-2\alpha -\min\Big\{\tfrac{2(1-1/K)}{N/K+1},\tfrac{1-K/N}{K+1}\Big\} \nonumber\\
    &\geq& 1-2\alpha -\sqrt{2/N} . \label{eq: KCV-CP is valid}
\end{IEEEeqnarray}
Therefore, validity for both cross-validation schemes is guaranteed for the larger miscoverage level of $2\alpha$. Accordingly, one can achieve miscoverage level of $\alpha$, satisfying \eqref{eq: set validity}, by considering the CV-CP set predictor $\Gamma^\text{CV}(x|\D)$ with $\alpha/2$ in lieu of $\alpha$ in \eqref{eq: prediction set KCV-CP classification}. That said, in the experiments, we will follow  the recommendation in \cite{barber2021predictive} and \cite{romano2020classification} to use $\alpha$ in  \eqref{eq: prediction set KCV-CP classification}.

\section{Online Conformal Prediction}\label{sec: Online Conformal Prediction}

In this section, we turn to online CP. Unlike the CP schemes presented in the previous section, online CP makes no assumptions about the probabilistic model underlying data generation \cite{gibbs2021adaptive, feldman2022conformalized}. Rather, it models the observations as a deterministic stream of input-output pairs $z[i]=(x[i],y[i])$ over time index $i=1,2,\dots$; and it targets a coverage condition defined in terms of the empirical rate at which the prediction set $\Gamma_i$ at time $i$ covers the correct output $y[i]$.

In the offline version of CP reviewed in the previous section, all $N$ samples of the data set $\D$ are assumed to be available upfront (see Fig.~\ref{fig: fig_tikz_cp_offline_online}(a)). In contrast, in online CP, a set predictor $\Gamma_i$ for time index $i$ is produced for each new input $x[i]$ over time $i=1,2,\dots$ Specifically, given the past observations $\{z[j]\}_{j=1}^{i-1}$, the set predictor $\Gamma_i\big(x[i]\big|\{z[j]\}_{j=1}^{i-1}\big)$ outputs a subset of the output space $\mathcal{Y}$. Given a target miscoverage level $\alpha~\in~[0,1]$, an online set predictor is said to be $(1-\alpha)$-\emph{long-term valid} if the following limit holds
\begin{equation}
    \lim_{I\to\infty} \frac{1}{I} \sum_{i=1}^I \indicator\Big( y[i] \in \Gamma_i\big(x[i]\big|\{z[j]\}_{j=1}^{i-1}\big) \Big)  = 1 - \alpha  \label{eq: long-term set validity}
\end{equation}
for all possible sequences $z[i]$ with $i=1,2,\dots$ Note that the condition \eqref{eq: long-term set validity}, unlike \eqref{eq: set validity}, does not involve any ensemble averaging with respect to the data distribution. We will take \eqref{eq: long-term set validity} as the relevant definition of calibration for online learning.

\emph{Rolling conformal inference} (RCI) \cite{feldman2022conformalized} adapts in an online fashion a \emph{calibration parameter} $\theta[i]$ across the time index $i$ as a function of the instantaneous error variable
\begin{equation}
    \mathrm{err}[i]=\indicator\big( y[i] ~\notin~ \Gamma_i(x[i]) \big),
\end{equation}
which equals $1$ if the correct output value is not included in the prediction set $\Gamma_i(x[i]) $, and $0$ otherwise. This is done using the update rule
\begin{equation}
    \theta[i+1] \gets \theta[i] + \gamma \big( \mathrm{err}[i] - \alpha\big), \label{eq: theta symmetric update}
\end{equation}
where $\gamma > 0$ is a learning rate. Accordingly, the parameter $\theta$ is increased by $\gamma(1-\alpha)$ if an error occurs at time $i$, and is decreased by $\gamma\alpha$ otherwise. Intuitively, a large positive parameter $\theta[i]$ indicates that the set predictor should be more inclusive in order to meet the validity constraint \eqref{eq: long-term set validity}; and vice versa, a large negative value of $\theta[i]$ suggests that the set predictor can reduce the size of the prediction sets without affecting the long-term validity constraint \eqref{eq: long-term set validity}.

Following \cite{feldman2022conformalized}, we elaborate on the use of the calibration parameter $\theta[i]$ in order to ensure condition \eqref{eq: long-term set validity} for an online version of the \naive quantile-based predictor \eqref{eq: naive quantile based} for scalar regression. A similar approach applies more broadly (see \cite{gibbs2021adaptive,zaffran2022adaptive}, and \cite{lin2022conformal}). Denote the data set $\D[i]=\{z[j]\}_{j=1}^{i-1}$ as having all previously observed labeled data set up till time $i-1$. The key idea behind RCI is to extend the \naive prediction interval \eqref{eq: naive quantile based} depending on the calibration parameter $\theta[i]$ as 
\begin{IEEEeqnarray}{rl}
    \label{eq: Gamma_i RCI y}
    &\Gamma_i^{\text{RCI}}\big(x[i]\big|\D[i]\big)  \\
    &= \Big[ \hat{y}(x|\phi_{\D[i],  \alpha/2}) - \varphi(\theta[i]),
             \hat{y}(x|\phi_{\D[i],1-\alpha/2}) + \varphi(\theta[i]) 
    \Big] , \nonumber
\end{IEEEeqnarray}
where
\begin{equation}
    \varphi(\theta)= \mathrm{sign}(\theta) \big(\exp^{|\theta|}-1\big) \label{eq: stretching func}
\end{equation}
is the so-called stretching function, a fixed monotonically increasing mapping. 

The set predictor RCI \eqref{eq: Gamma_i RCI y} ``corrects'' the NQB set predictor \eqref{eq: naive quantile based}, via the additive stretching function $\varphi(\theta[i])$ based on the calibration parameter $\theta[i]$. As the time index $i$ rolls, the calibration parameter $\theta[i]$ adaptively inflates and deflates according to \eqref{eq: theta symmetric update}. Upon each observation of new label $y[i]$, the quantile predictor model parameters $\phi_{\D[i],\alpha/2}$ and $\phi_{\D[i],1-\alpha/2}$ can also be updated, without affecting the long-term validity condition \eqref{eq: long-term set validity} \cite[Theorem 1]{feldman2022conformalized}. We refer to Appendix~\ref{appendix: Algorithmic Details for Rolling Conformal Inference} for further details on online CP.

\section{Symbol Demodulation}\label{sec: Symbol Demodulation}

In this section, we focus on the application of offline CP, as described in Sec.~\ref{sec: Conformal Prediction}, to the problem of symbol demodulation in the presence of transmitter hardware imperfections. This problem was also considered in \cite{park2021fewpilots,cohen2022bayesian} by focusing on frequentist and Bayesian learning. Unlike \cite{park2021fewpilots,cohen2022bayesian}, we investigate the use of CP as a means to obtain set predictors satisfying the validity condition \eqref{eq: set validity}.

\subsection{Problem Formulation}

The problem of interest consists of the demodulation of symbols from a discrete constellation based on received baseband signals subject to hardware imperfections, noise, and fading. The goal is to design \emph{set demodulators} that output a subset of all possible constellation points with the guarantee that the subset includes the true transmitted signal with the desired target probability $1-\alpha$. In the context of channel decoding, this type of receiver is referred to as a list decoder \cite{tal2015list}.

To keep the notation consistent with the previous sections, we write as $y[i]$ the $i$-th transmitted symbols, and as $x[i]$ the corresponding received signal. Each transmitted symbol $y[i]$ is drawn uniformly at random from a given constellation $\mathcal{Y}$.
 We model I/Q imbalance at the transmitter and phase fading as in \cite{cohen2023calibrating}. Accordingly, the ground-truth channel law connecting symbols $y[i]$ into received samples $x[i]$ is described by the equality
\begin{equation}
    \rv{x}[i] = e^{\jmath \rvpsi} f_{\text{IQ}}(\rv{y}[i]) + \rv{v}[i], \label{eq: demodulation channel model}
\end{equation}
for a random phase $\rvpsi \sim \mathrm{U} [0,2\pi)$, where the additive noise is $ \rv{v}[i] \sim \mathcal{CN}(0,\SNR^{-1}) $ for signal-to-noise ratio level $\SNR$. Furthermore, the I/Q imbalance function \cite{tandur2007joint} is defined as
\begin{equation}
    f_{\text{IQ}}(\rv{y}[i]) 
    = \bar{\rv{y}}_{\text{I}}[i] + \jmath \bar{\rv{y}}_{\text{Q}}[i] , \label{eq:iq_imbalance}
\end{equation}
where
\begin{equation}
    \begin{bmatrix} 
        \bar{\rv{y}}_{\text{I}}[i] \\ \bar{\rv{y}}_{\text{Q}}[i]
    \end{bmatrix}
    =
    \begin{bmatrix} 
        1+\rvepsilon & 0 \\ 0 & 1-\rvepsilon
    \end{bmatrix}
    \begin{bmatrix} 
        \cos \rvdelta & -\sin \rvdelta \\ -\sin \rvdelta & \cos \rvdelta
    \end{bmatrix}
    \begin{bmatrix} 
        \rv{y}_{\text{I}}[i] \\ \rv{y}_{\text{Q}}[i]
    \end{bmatrix}, \label{eq: iq imbalance as matrix}
\end{equation}
with $\rv{y}_{\text{I}}[i] $ and $\rv{y}_{\text{Q}}[i]$ being the real and imaginary parts of the modulated symbol $\rv{y}[i]$; and  $\bar{\rv{y}}_{\text{I}}[i]$  and $\bar{\rv{y}}_{\text{Q}}[i]$ standing for the real and imaginary parts of the transmitted symbol $f_{\text{IQ}}(\rv{y}[i])$. In \eqref{eq: iq imbalance as matrix}, the channel state $\rv{c}$ consists of the tuple $\rv{c}=(\rvpsi,\rvepsilon,\rvdelta)$ encompassing the complex phase $\rvpsi$ and the I/Q imbalance parameters $(\rvepsilon,\rvdelta)$. 

\subsection{Implementation}

As in \cite{park2021fewpilots,cohen2022bayesian}, demodulation is implemented via a neural network probabilistic model $p(y|x,\phi)$ consisting of a fully connected network with real inputs $x[i]$ of dimension $2$ as per \eqref{eq: demodulation channel model}, followed by three hidden layers with $10,30$, and $30$ neurons having ReLU activations in each layer. The last layer implements a softmax classification for the $|\mathcal{Y}|$ possible constellation points.

We adopt the standard NC score \eqref{eq: NC classification}, where the trained model $\phi_\D$ for frequentist learning is obtained via $I=120$ GD update steps for the minimization of the cross-entropy training loss with learning rate $\eta=0.2$; while for Bayesian learning we implement a gradient-based MC method, namely Langevin MC, with burn-in period of  $R_\text{min}=100$, ensemble size $R=20$, learning rate $\eta=0.2$, and temperature parameter $T=20$. We assume standard Gaussian distribution for the prior distribution \cite{welling2011bayesian}. Details on Langevin MC can be found in Appendix~\ref{appendix: Frequentist and Bayesian Learning}.

We compare the \naive set predictor \eqref{eq: prediction set naive}, also studied in \cite{park2021fewpilots, cohen2022bayesian}, which provides no formal coverage guarantees, with the CP set prediction methods reviewed in Sec.~\ref{sec: Conformal Prediction}. VB-CP uses equal set sizes for the training and validation sets. We target the miscoverage level as $\alpha=0.1$.

\subsection{Results}

We consider the Amplitude-Phase-Shift-Keying (APSK) modulation with $|\mathcal{Y}|=8$. The SNR level is set to $\SNR=5\text{ dB}$. The amplitude and phase imbalance parameters are independent and distributed as $\rvepsilon \sim \Betadist(\epsilon / 0.15 | 5,2)$ and $\rvdelta \sim \Betadist(\delta / 15^{\circ} | 5,2)$, respectively \cite{park2021fewpilots}.

\begin{figure}
    \centering
    \includegraphics[trim=3.5cm 9cm 4cm 10cm, clip, width=8cm]{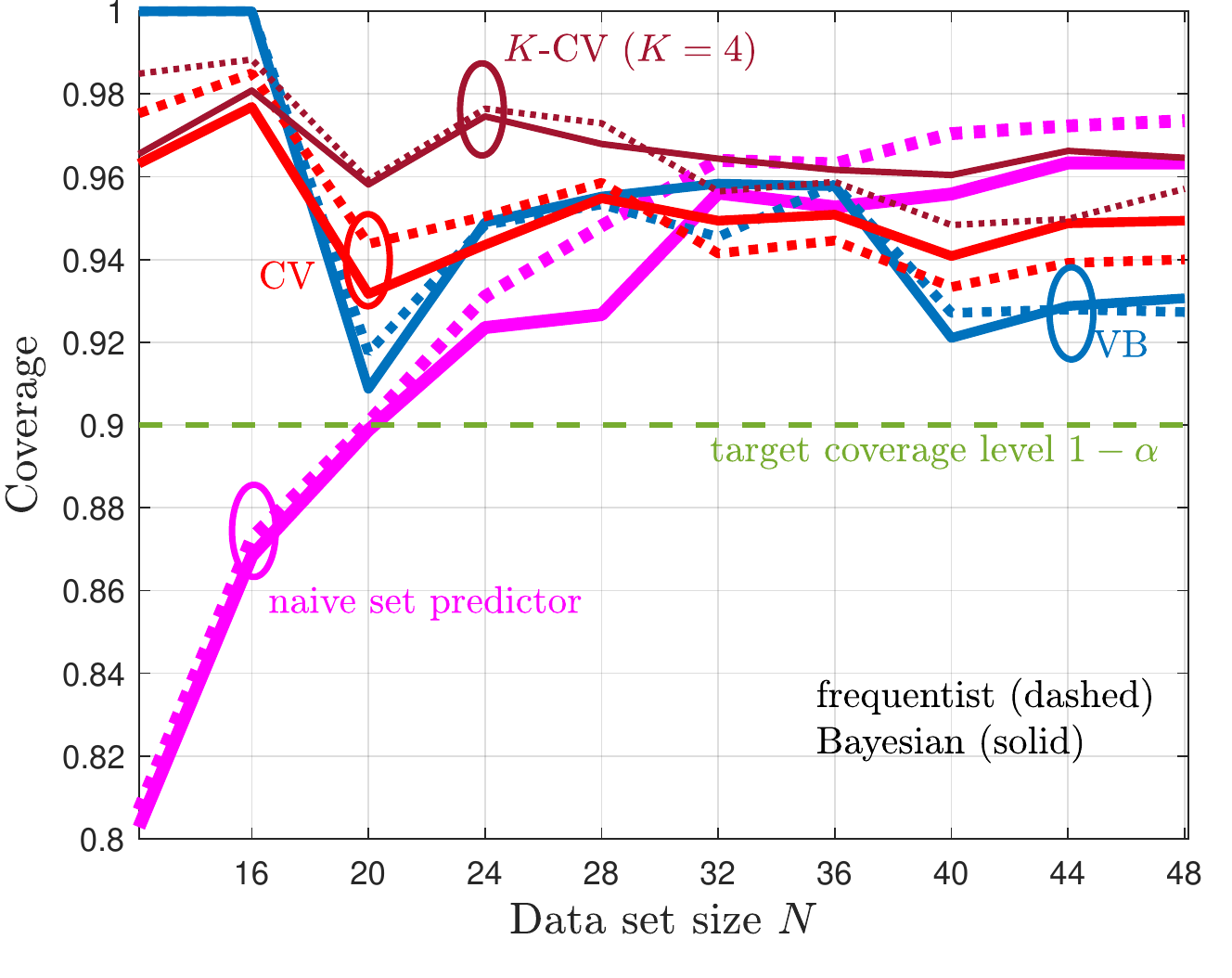}
    \caption{Coverage for \naive set predictor \eqref{eq: prediction set naive}, VB-CP \eqref{eq: prediction set VB}, CV-CP, and $K$-CV-CP \eqref{eq: prediction set KCV-CP classification} with $K=4$, for symbol demodulation problem 
    (Section \ref{sec: Symbol Demodulation}). For every set predictors, the NC scores are evaluated either using frequentist learning (dashed lines) or Bayesian learning (solid lines). The coverage level is set to $1-\alpha=0.9$, and each numerical evaluation is averaged over $50$ independent trials (new channel state $c$) with $\Nte=100$ test points.}
    \label{fig: coverage_vs_N_demod}
    \vspace{0.5cm}
    \includegraphics[trim=3.5cm 9cm 4cm 10cm, clip, width=8cm]{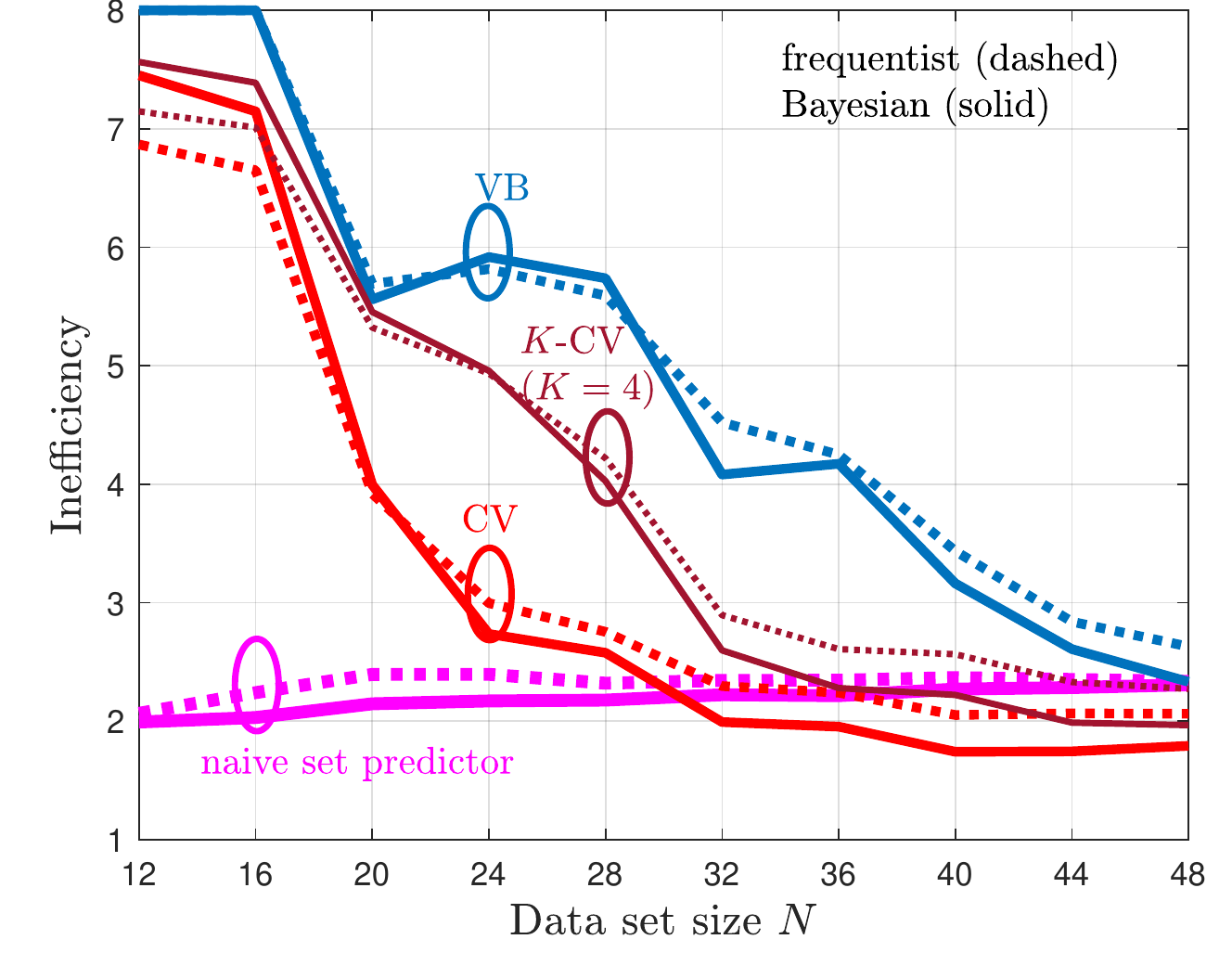}
    \caption{Average set prediction size (inefficiency) for the same setting of Fig.~\ref{fig: coverage_vs_N_demod}.}
    \label{fig: ineff_vs_N_demod}
\end{figure}
Fig.~\ref{fig: coverage_vs_N_demod} shows the 
\begin{equation} 
    \text{empirical coverage} = 
    \tfrac{1}{\Nte} \sum_{j=1}^\Nte \indicator \big( y^\text{te}[j] \in \Gamma(x^\text{te}[j]|\D) \big) , \label{eq: empirical coverage}
\end{equation}
and Fig.~\ref{fig: ineff_vs_N_demod} shows the
\begin{equation}
    \text{empirical inefficiency} = 
    \tfrac{1}{\Nte} \sum_{j=1}^\Nte  \big| \Gamma(x^\text{te}[j]|\D) \big| , \label{eq: empirical inefficiency}
\end{equation}
both evaluated on a test set $\Dte=\{(x^\text{te}[j],y^\text{te}[j])\}_{j=1}^{\Nte}$ with $\Nte=100$, as a function of the size of the available data set $\D$. We average the results for 50 independent trials, each corresponding to independent draws of the variables $\{\D,\Dte\}$ from the ground truth distribution.  This way, the metrics \eqref{eq: empirical coverage}-\eqref{eq: empirical inefficiency} provide an estimate of the coverage \eqref{eq: set validity} and of the inefficiency \eqref{eq: ineff(Gamma) = E | Gamma |}, respectively \cite{barber2021predictive}.

From Fig.~\ref{fig: coverage_vs_N_demod}, we first observe that the \naive set predictor, with both frequentist and Bayesian learning, does not meet the desired coverage level in the regime of a small number $N$ of available samples. In contrast, confirming the theoretical calibration guarantees presented in Sec.~\ref{sec: Conformal Prediction}, all CP methods provide coverage guarantees, achieving coverage rates above $1-\alpha$. Furthermore, as seen in Fig.~\ref{fig: ineff_vs_N_demod}, coverage guarantees are achieved by suitably increasing the size of prediction sets, which is reflected by the larger inefficiency. The size of the prediction sets, and hence the inefficiency, decreases as the data set size, $N$, increases. In this regard, due to their more efficient use of the available data, CV-CP and $K$-CV-CP predictors have a lower inefficiency as compared to VB predictors, with CV-CP offering the best performance. Finally, Bayesian NC scores are generally seen to yield set predictors with lower inefficiency, confirming the merits of Bayesian learning in terms of calibration.

\section{Modulation Classification}\label{sec: Modulation Classification}

In this section, we propose and evaluate the application of offline CP to the problem of modulation classification \cite{o2016convolutional, o2018over}.

\subsection{Problem Formulation}

Due to the scarcity of frequency bands, electromagnetic spectrum sharing among licensed and unlicensed users is of special interest to improve the efficiency of spectrum utilization. In sensing-based spectrum sharing, a transmitter scans the prospective frequency bands to identify, for each band, if the spectrum is occupied, and, if so, if the signal is from a licensed user or not. A key enabler for this operation is the ability to classify the modulation of the received signal \cite{zhu2015automatic}. The modulation classification task is made challenging by the dimensionality of the baseband input signal and by the distortions caused by the propagation channel. Data-driven solutions \cite{zhou2020deep} have shown to be effective for this problem in terms of accuracy, while the focus here is on calibration performance.

Accordingly, we aim at designing \emph{set modulation classifiers} that output a subset of the set of all possible modulation schemes with the property that the true modulation scheme is contained in the subset with a desired probability level $1-\alpha$.  To this end, we adopt the data set provided by \cite{oshea2018over}, which has approximately $2.5\times10^6$ baseband signals of $1024$ I/Q samples, each produced using one out of $24$ possible digital and analog modulations across different SNR values and channel models. We focus only on the high SNR regime ($\geq 6$ dB). This data set $\D$ is made out of approximately $1.28\times 10^6$ $(x,y)$ pairs, where $x$ is the channel output signal of dimension $2048$ and $y$ is the index of one of the $|\mathcal{Y}|=24$ possible modulations. The SNR value itself is not available to the classifier.

\subsection{Implementation}\label{sec: Modulation Classification: Implementation}

We use a neural network architecture similar to the one used in \cite{oshea2018over}, which has $7$ one-dimensional convolutional layers with kernel size $3$ and $64$ channels for all layers, except for the first layer with has $2$ channels. The convolution layers are followed by $3$ fully-connected linear layers. A scaled exponential linear unit (SELU) is used for all inner layers, and a softmax is used at the last, fully connected, layer. We assume availability of $N=4800$ pairs $(x,y)$ for the data set $\D$, while gauging the empirical inefficiency and coverage level with $\Nte=1000$ held-out pairs. A total number of $I=4000$ GD steps with fixed learning rate of $0.02$ are carried out, and the target miscoverage rate is set to $\alpha=0.1$. VB partitions its available data into equal sets for training and validation.

\subsection{Results}
\begin{figure}
    \centering
    \includegraphics[page=1, trim=0.3cm 0.3cm 0.3cm 0.3cm, clip, width=4.2cm]{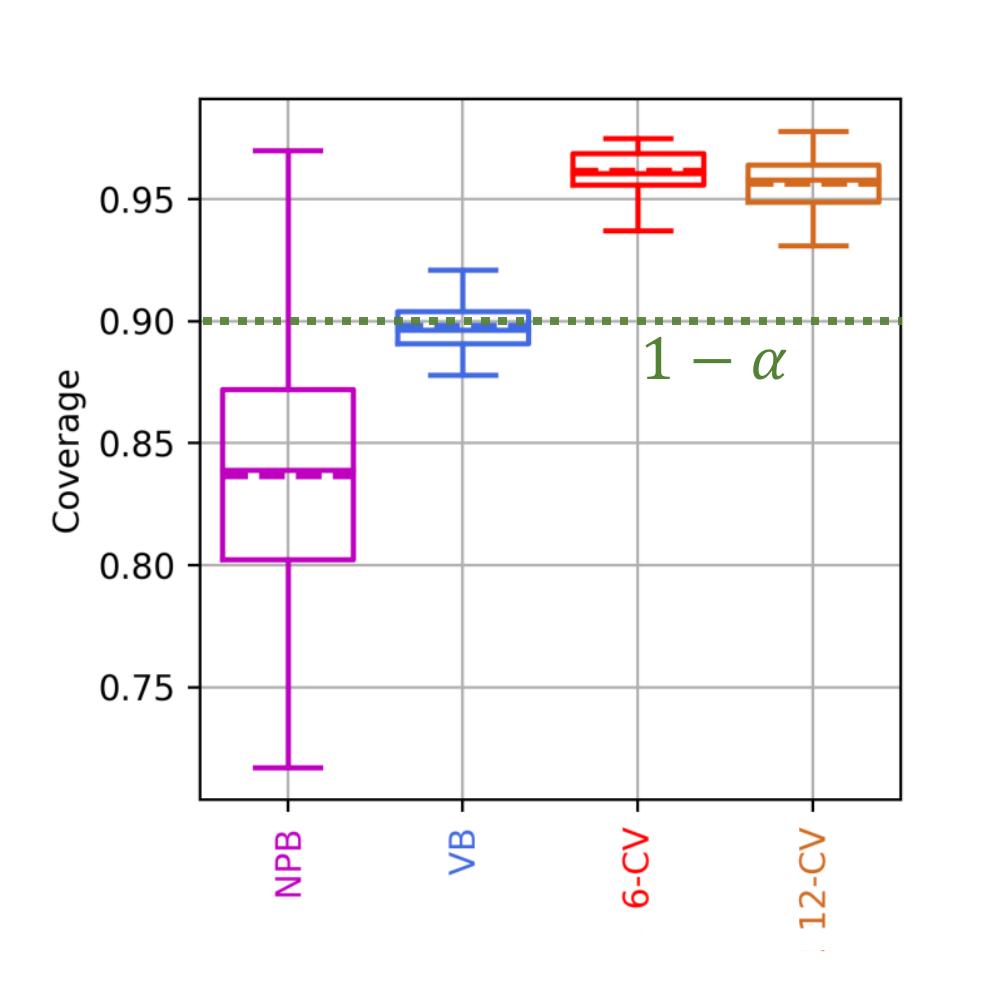}
    \includegraphics[page=2, trim=0.3cm 0.3cm 0.3cm 0.3cm, clip, width=4.2cm]{Figs/fig_coverg_ineff_2022_11_16__16_20.pdf}
    \caption{Coverage and inefficiency for NPB \eqref{eq: prediction set naive}, VB-CP \eqref{eq: prediction set VB}, and $K$-CV-CP \eqref{eq: prediction set KCV-CP classification} with $K=6$ and $K=12$, for the modulation classification problem (implementation details in Section \ref{sec: Modulation Classification: Implementation}). 
    The boxes represent the $25\%$ (lower edge), $50\%$ (solid line within the box), and $75\%$ (upper edge) percentiles of the empirical performance metrics evaluated over $32$ different experiments, with average value shown by the dashed line.}
    \label{fig: mod_class_plotbox}
\end{figure}

In this problem, due to computational cost, we exclude CV-CP and we focus on $K$-CV-CP with a moderate number of folds, namely $K=6$ and $K=12$. In Fig.~\ref{fig: mod_class_plotbox}, box plots show the quartiles of the empirical coverage \eqref{eq: empirical coverage} and of the empirical inefficiency \eqref{eq: empirical inefficiency} from $32$ independent runs, with different realizations of data set and test examples. The lower edge of the box represents the $0.25$-quantile; the solid line within the box the median; the dashed line within the box the average; and the upper edge of the box the $0.75$-quantile. As can be seen in the figure, the \naive set predictor is invalid (see average shown as dashed line), and it exhibits a wide spread of the coverage rates across the trials. On the other hand, all CP set predictors are valid, meeting the predetermined coverage level $1-\alpha=0.9$, and have less spread-out coverage rates. 

As also noted in the previous section, VB-CP suffers from larger predicted set size as compared to $K$-CV-CP, due to poor sample efficiency. A small number of folds, as low as $K=6$, is sufficient for $K$-CV-CP to outperform VB-CP. This improvement in efficiency comes at the computational cost of training six models, as compared to the single model trained by VB-CP.

\section{Online Channel Prediction}\label{sec: Online Channel Prediction}

In this section, we investigate the use of online CP, as described in Sec.~\ref{sec: Online Conformal Prediction}, for the problem of channel prediction. We specifically focus on the prediction of the received signal strength (RSS), which is a key primitive at the physical layer, supporting important functionalities such as resource allocation \cite{wong2009optimal, nikoloska2022modular}.

\subsection{Problem Formulation}
Consider a receiver that has access to a sequence of RSS samples from a given device. 
We aim at designing a predictor that, given a sequence of past samples from the RSS sequence,  produces an \emph{interval} of values for the next RSS sample. To meet calibration requirements, the interval must contain the correct future RSS value with the desired rate level $1-\alpha$. Unlike the previous applications, here the rate of coverage is evaluated based on the time average \begin{equation}
     \frac{1}{t}\sum_{i=1}^t \indicator\Big( y[i] \in \Gamma_i\big(x[i]\big|\{z[j]\}_{j=1}^{i-1}\big) \Big).    \label{eq: long-term set validity1}
\end{equation} This is computed as the fraction previous time instants $i\in \{1,...,t\}$ at which the set predictor $\Gamma_i$ includes the true RSS value $y[i]$.

We consider two data sets of RSS sequences. The first data set records RSS samples $y[i]$ in logarithmic scale for an IEEE 802.15.4 radio over time index $i$ \cite{zanella2013rss}. We further use the available side information on the time-variant channel ID, which determines the carrier frequency used at time $i$ out of the $16$ possible bands, as the input $x[i]$. At time $i$, we observe a sequence of RSS samples $z[1],\dots,z[i-1]$ with $z[i]=(x[i],y[i])$, and the goal is to predict the next RSS sample $y[i]$ via the online set predictor $\Gamma_i^{\text{RCI}}$ \eqref{eq: Gamma_i RCI y}. 

The second data set \cite{simmons2022ai} reports samples $y[i]$, measured in dBm, on a $5.8$ GHz device-to-device link without additional input. Hence, in this case, we predict the next RSS sample $y[i]$ using the previous RSS samples $y[1],\ldots,y[i-1]$. Note that the prior works \cite{zanella2013rss,simmons2022ai} adopted standard probabilistic predictors, while here we focus on set predictors that produce a prediction interval $\Gamma_i^{\text{RCI}}\big(x[i]\big|\{z[j]\}_{j=1}^{i-1}\big)$.

\subsection{Implementation}
We build the CP set predictor by leveraging the probabilistic neural network used in \cite{feldman2022conformalized} as the model class for the quantile predictors in \eqref{eq: phi_lo and phi_hi as pinball minimizers}-\eqref{eq: naive quantile based}. Each quantile predictor consists of a multi-layer neural network that pre-processes the most recent $K$ pairs $\{z[i-K],\dots,z[i-1]\}$; of a stacked long short-term memory (LSTM) \cite{hochreiter1997long} with two layers; and of a post-processing neural network, which maps the last LSTM hidden vector into a scalar that estimates the quantile used in \eqref{eq: naive quantile based}. For details of the implementation, we refer to Appendix~\ref{appendix: Implementation of Online Channel Prediction}.

\subsection{Results}

\begin{figure}
    \centering
    \includegraphics[trim=0.5cm 0.0cm 1.2cm 0.5cm, clip, width=7cm]{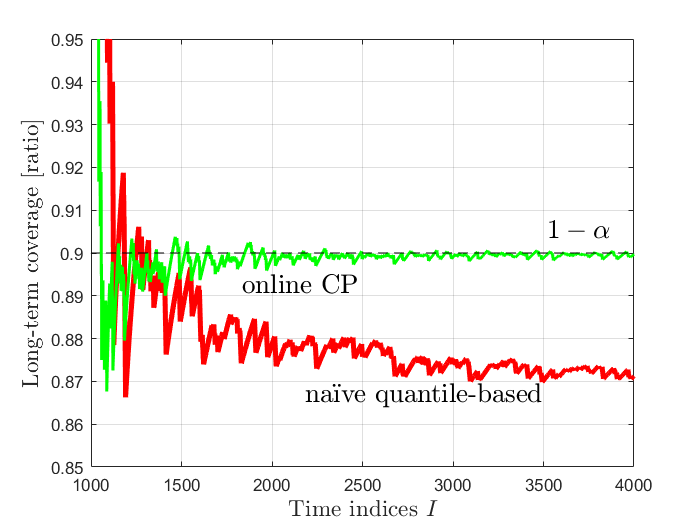}
    \includegraphics[trim=0.5cm 0.0cm 1.2cm 0.5cm, clip, width=7cm]{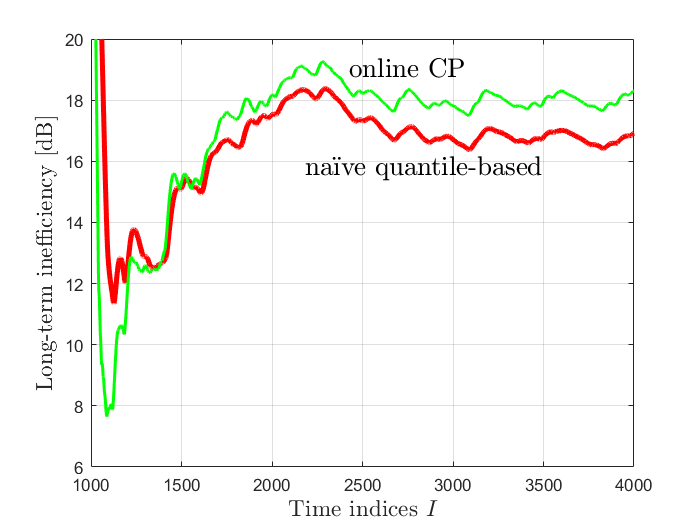}
    \caption{CP for time-series \texttt{Outdoor} of \cite{zanella2013rss}: \textbf{(top)} Time-average coverage \eqref{eq: time-average coverage} of \naive set prediction and online CP; \textbf{(bottom)} Time-averaged inefficiency \eqref{eq: time-average inefficiency} of \naive set prediction and online CP.}
    \label{fig: Zanella results}
\end{figure}

\begin{figure}
    \centering
    \includegraphics[trim=0.5cm 0.0cm 1.2cm 0.5cm, clip, width=7cm]{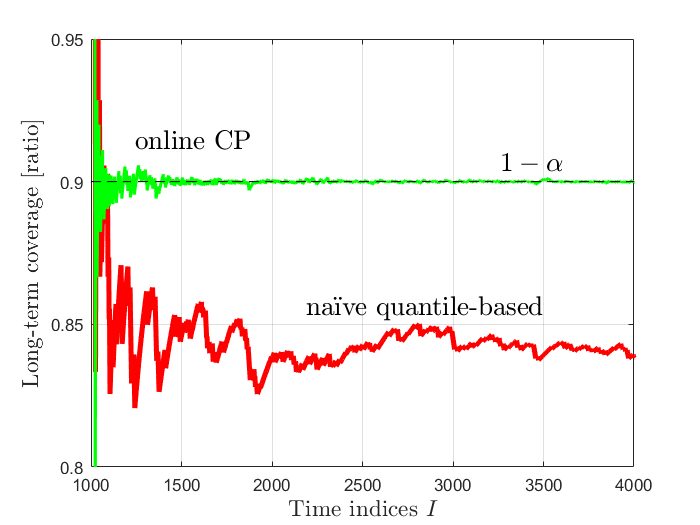}
    \includegraphics[trim=0.5cm 0.0cm 1.2cm 0.5cm, clip, width=7cm]{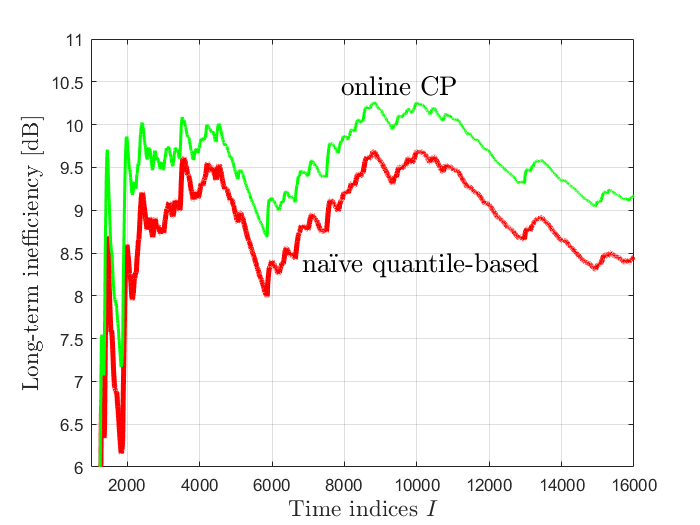}
    \caption{CP for time-series \texttt{NLOS\_Head\_Indoor\_1khz} in \cite{simmons2022ai}: \textbf{(top)} Time-average coverage \eqref{eq: time-average coverage} of \naive set prediction and online CP; \textbf{(bottom)} Time-average inefficiency \eqref{eq: time-average inefficiency} of \naive  set prediction and online CP.}
    \label{fig: Simmons results}
\end{figure}

Fig.~\ref{fig: Zanella results} and Fig.~\ref{fig: Simmons results} report the 
\begin{equation}
    \text{time-average coverage} = \tfrac{1}{I} \sum_{i=1}^I \indicator\Big( y[i] \in \Gamma_i\big(x[i]\big|\{z[j]\}_{j=1}^{i-1}\big) \Big) \label{eq: time-average coverage}
\end{equation}
and the
\begin{equation}
    \text{time-average inefficiency} = \tfrac{1}{I} \sum_{i=1}^I \Big| \Gamma_i\big(x[i]\big|\{z[j]\}_{j=1}^{i-1}\big) \Big| \label{eq: time-average inefficiency}
\end{equation}
for online CP \eqref{eq: Gamma_i RCI y}, compared to a baseline of the \naive quantile-based predictor \eqref{eq: naive quantile based}, as a function of the time window size $I$ for data sets \cite{simmons2022ai} and \cite{simmons2022ai}, respectively. We have discarded $1000$ samples for a warm-up period for both metrics \eqref{eq: time-average coverage} and \eqref{eq: time-average inefficiency}. 

In both cases, the \naive predictor is seen to fail to satisfy the coverage condition \eqref{eq: long-term set validity} for both data sets, while online CP converges to the target level $1-\alpha=0.9$. This result is obtained by online CP with a modest increase of around $8\%$ for both data sets in terms of inefficiency.

\section{Conclusions}\label{sec: conclusions}

AI in communication engineering should not only target accuracy, but also calibration, ensuring a reliable and safe adoption of machine learning within the overall telecommunication ecosystem. In this paper, we have proposed the adoption of a general framework, known as conformal prediction (CP), to \emph{transform} any existing AI model into a well-calibrated model via post-hoc calibration for communication engineering. Depending on the situation of interest, post-hoc calibration leverages either an held-out (cross) validation set or previous samples. Unlike calibration approaches that do not formally guarantee reliability, such as Bayesian learning or temperature scaling, CP provides formal guarantees of calibration, defined either in terms of ensemble averages or long-term time averages. Calibration is retained irrespective of the accuracy of the trained models, with more accurate models producing smaller set predictions.

To validate the reliability of CP-based set predictors, we have provided extensive comparisons with conventional methods based on Bayesian or frequentist learning. Focusing on demodulation, modulation classification, and channel prediction, we have demonstrated that AI models calibrated by CP provide formal guarantees of reliability, which are practically essential to ensure calibration in the regime of limited data availability.

Future work may consider applications of CP to other use cases in wireless communication systems, as well as extension of involving training-based calibration \cite{kumar2018trainable, einbinder2022training} and/or meta-learning \cite{park2022few}.

\appendices

\section{Frequentist and Bayesian Learning}\label{appendix: Frequentist and Bayesian Learning}
Given access to training data set $\D=\big\{z[i]=(x[i],y[i])\big\}_{i=1}^N$ with $N$ examples, frequentist learning finds a model parameter vector $\phi_{\D}^*$ by tackling the following empirical risk minimization (ERM) problem
\begin{IEEEeqnarray}{rcl}
    \min_\phi \Big\{ L_\D(\phi) &=& - \tfrac{1}{N} \sum_{(x,y)\in\D} \log p(y|x,\phi) \label{eq: L_D empirical loss}\\
    &=& \E_{p_\D(x,y)} \big[ -\log p(\rv{y}|\rv{x},\phi) \big] \Big\}, \nonumber
\end{IEEEeqnarray}
with empirical distribution $p_\D(x,y)$ defined by the data set $\D$.

Bayesian learning addresses epistemic uncertainty by treating the model parameter vector as a random vector $\rvphi$ with prior distribution $\rvphi \sim p(\phi)$.
Ideally, Bayesian learning updates the prior $p(\phi)$ to produce the posterior distribution $p(\phi|\D)$ as
\begin{equation}
    p(\phi|\D) \propto p(\phi) \prod_{i=1}^N p\big(y[i]\big|x[i],\phi\big) \label{eq: p(phi|D)}
\end{equation}
and obtains the ensemble predictor for the test point $(x,y)$ by averaging over multiple models, i.e., 
\begin{equation}
    p(y|x,\D) = \E_{p(\phi|\D)} [p(y|x,\rvphi)]. 
    \label{eq: Bayes pred = E_p [p]}
\end{equation}

In practice, as the true posterior distribution is generally intractable due to the normalizing factor in \eqref{eq: p(phi|D)}, approximate Bayesian approaches are considered via VI or MC techniques (see, e.g., \cite{simeone2022machine}).

In the experiments, we adopted Langevin MC  to approximate the Bayesian posterior \cite{welling2011bayesian,simeone2022machine}. Langevin MC adds  Gaussian noise to each standard GD update for frequentist learning (see, e.g., \cite[Sec.~4.10]{simeone2022machine}). The noise has power $2\eta/T$, where $\eta$ is the GD learning rate and $T>0$ is a temperature parameter.  Langevin MC produces $R$ model parameters $\{\phi[r]\}_{r=1}^R$ across $R$ consecutive iterations. We specifically retain only the last $R$ samples, discarding an initial  burn-in period of $R_\text{min}$ iterations. The temperature parameter $T$ is typically chosen to be larger than $1$ \cite{wenzel2020good, ye2017langevin}. With the $R$ samples, the expectation term in \eqref{eq: Bayes pred = E_p [p]} is approximated as the empirical average $\frac{1}{R}\sum_{r=1}^R p(y|x,\phi[r])$.

We observe that Langevin MC is a probabilistic training algorithm, and that it satisfies the permutation-invariance property in terms of the distribution of the random output models discussed in Sec.~\ref{sec: Calibration Guarantees}.

\section{Algorithmic Details for Rolling Conformal Inference}\label{appendix: Algorithmic Details for Rolling Conformal Inference}

The RCI algorithm is reproduced from \cite{feldman2022conformalized} in Algorithm~\ref{alg: Rolling Conformal Inference for Regression}.

\begin{algorithm}
    \caption{\texttt{Rolling Conformal Inference} (for Regression) \cite{feldman2022conformalized}}
    \label{alg: Rolling Conformal Inference for Regression}
    \KwInputs{      $\alpha=$ long-term target miscoverage level\\
                    $\theta[1]=$ initial calibration parameter\\
                    $\phi^\text{lo}[1],\phi^\text{hi}[1]=$ initial models}
    \KwParameters{  $I=$ number of online iterations\\
                    $\gamma=$ learning rate for calibration parameter\\
                    $\eta=$ learning rate for model updates\\
                    }
    \KwOutput{      $\{\Gamma^\text{RCI}_{i}\big(x[i]\big|\{z[j]\}_{j=1}^{i-1}\big)\}_{i=1}^I$ = predicted sets for $\{x[i]\}_{i=1}^I$}
    \BlankLine
    \For{$i=1,\dots,I$ time instants}{
        Retrieve a new data sample $(x[i],y[i])$\;
        \algcomment{Set prediction of new input}
        Calculate set $\Gamma^\text{RCI}_{i}\big(x[i]\big|\{z[j]\}_{j=1}^{i-1}\big)$ using\; 
        $\Big[ \hat{y}\big(x[i]\big|\phi^\text{lo}[i]\big) - \varphi(\theta[i]),
               \hat{y}\big(x[i]\big|\phi^\text{hi}[i]\big) + \varphi(\theta[i]) \Big]$\;
        \algcomment{Check if prediction is unsuccessful}
        $\mathrm{err}[i] \gets \indicator\big(y[i] \notin \Gamma^\text{RCI}_{i}\big(x[i]\big|\{z[j]\}_{j=1}^{i-1}\big) \big)$\;
        \algcomment{Update calibration parameter}
        $\theta[i+1] \gets \theta[i] + \gamma \big( \mathrm{err}[i]  - \alpha\big) $\; \label{algline: update theta}
        \algcomment{Update models using new sample}
        $\phi^\text{lo}[i+1]\gets \phi^\text{lo}[i] - \eta \nabla_{\phi} \ell_{\alpha/2} \Big( y[i] , \hat{y}\big(x[i]\big|\phi^\text{lo}[i]\big) \Big)$\;
        $\phi^\text{hi}[i+1]\gets \phi^\text{hi}[i] - \eta \nabla_{\phi} \ell_{1-\alpha/2} \Big( y[i] , \hat{y}\big(x[i]\big|\phi^\text{hi}[i]\big) \Big)$\;
        }
    \Return predicted sets $\{\Gamma^\text{RCI}_{i}\big(x[i]\big|\{z[j]\}_{j=1}^{i-1}\big)\}_{i=1}^I$
\end{algorithm}

\section{Implementation of Online Channel Prediction}\label{appendix: Implementation of Online Channel Prediction}
The architecture of the set predictor is inspired by \cite{feldman2022conformalized}, 
and made out of three artificial neural networks. The first, $f_\text{pre}(\cdot)$, is a multi-layer perceptron (MLP) network with hidden layers of $16,32$ neurons each, parametrized by vector $\phi_\text{pre}[i]$. It is meant to apply a pre-process over the most recent observed $K=20$ pairs $\{z[i-K],\dots,z[i-1]\}$ to be transformed element-wise into a length-$K$ vector $w[i]=\big[w_1[i],\dots,w_K[i]\big]^\top$, in which the $k$-th element ($k=1,\dots,K$) is $w_k[i]= f_\text{pre}\big(z[i-K+k-1]\big|\phi_\text{pre}[i]\big)$. Effectively, this will serve as a temporal sliding $K$-length window, with a time-evolving pre-processing function. The second neural network, $f_\text{LSTM}(\cdot)$ has two layers with model parameter vectors $\phi_\text{LSTM}^1[i]$ (first layer) and $\phi_\text{LSTM}^2[i]$ (second layer), which retains a memory via the hidden state vectors $h$ and $c$, initialized at every time index $i$ as $c_0^1[i]=c_0^2[i]=h_0^1[i]=h_0^2[i]=\rv{0}$. By accessing the previous $K$ pairs via the vector $w[i]$, this recurrent neural network extracts temporal patterns by sequentially transferring information via LSTM cells (with shared parameter vectors) in the image of hidden and cell state vectors $c_{k}[i], h_{k}[i]$ via the LSTM cells. These vectors flow along the LSTM by concatenating $k=1,\dots,K$ cells, and forming vectors of length $32$ each
\begin{align}
    [c_k^1[i],h_k^1[i]] &=& f_\text{LSTM}\big(w_k  [i],c_{k-1}^1[i],h_{k-1}^1[i]\big|\phi_\text{LSTM}^1[i]\big) \\
    [c_k^2[i],h_k^2[i]] &=& f_\text{LSTM}\big(h_k^1[i],c_{k-1}^2[i],h_{k-1}^2[i]\big|\phi_\text{LSTM}^2[i]\big) .
\end{align}
The third and last network is a post-processing MLP $f_\text{post}(\cdot)$ with one hidden layer of $32$ neurons, and with parameter vector $\phi_\text{post}[i]$, which maps the last LSTM hidden $64$-length vector $h_K[i]=[h_K^1[i],h_K^2[i]]^\top$ into a scalar $f_\text{post}\big(x[i],h_K[i]\big|\phi_\text{post}[i]\big) \in \mathbb{R}$ that estimates the quantile for the output $y[i]$. Accordingly, the time evolving model parameter is the tuple 
\begin{equation}
    \phi[i]=(\phi_\text{pre}[i],\phi_\text{LSTM}^1[i],\phi_\text{LSTM}^2[i],\phi_\text{post}[i]).
\end{equation}
This model is instantiated twice for the regression problem: one for the $\alpha/2$ lower quantile and the other for $1-\alpha/2$ upper quantile. For every time instant $i$, after the new output $y[i]$ is observed, continual learning of the models is taken place by training the models with corresponding pinball losses \eqref{eq: def pinball loss} using the new pair $(x[i],y[i])$, while initializing the models as the previous models at time instant $i-1$. 

The miscoverage rate was set to $\alpha=0.1$, the learning rate to $\eta=0.01$, and we chose $\gamma=0.03$ for the calibration parameter $\theta$ in \eqref{eq: theta symmetric update}.


\vspace{-0.15cm}
\bibliographystyle{IEEEtran} 
\bibliography{my_bib.bib} 

\vspace{1.5cm}

\vfill

\end{document}